\documentclass[journal]{IEEEtran}

\usepackage{graphicx}
\usepackage{amsmath}
\usepackage{multirow}
\usepackage{mathtools}
\usepackage{placeins}
\usepackage{float}
\usepackage{booktabs}
\usepackage{ragged2e}
\usepackage{cite}

\usepackage{amsfonts}
\usepackage{xcolor}
\usepackage{romannum}
\usepackage{bm}

\usepackage{graphicx}
\usepackage{amsmath}
\usepackage{multirow}
\usepackage{mathtools}
\usepackage{placeins}
\usepackage{float}
\usepackage{booktabs}
\usepackage{ragged2e}
\usepackage{amssymb}
\usepackage{stackengine}
\usepackage{subcaption}
\usepackage{footnote}

\usepackage{bm}
\usepackage{subcaption}
\usepackage{rotating}
\usepackage{hyperref}
\usepackage{cite}

\usepackage{stackengine}
\usepackage{stmaryrd}
\ifCLASSINFOpdf
\else
\fi

\begin{document}

\title{Partial sequence labeling with structured Gaussian Processes}

\author{Xiaolei~Lu, Tommy~W.S. Chow,~\IEEEmembership{Fellow,~IEEE}%

\IEEEcompsocitemizethanks{\IEEEcompsocthanksitem Xiaolei Lu is with
the Dept of Electrical Engineering at the City University of Hong
Kong, Hong Kong (Email:xiaoleilu2-c@my.cityu.edu.hk)
\IEEEcompsocthanksitem Tommy W S Chow is with the Dept of Electrical
Engineering at the City University of Hong Kong, Hong Kong (Corresponding author: E-mail,
eetchow@cityu.edu.hk; contact: (852) 2788 7756).}}% <-this % stops an unwanted space

\maketitle

\begin{abstract}
Existing partial sequence labeling models mainly focus on max-margin framework which fails to provide an uncertainty estimation of the prediction. Further, the unique ground truth disambiguation strategy employed by these models may include wrong label information for parameter learning. In this paper, we propose structured Gaussian Processes for partial sequence labeling (SGPPSL), which encodes uncertainty in the prediction and does not need extra effort for model selection and hyperparameter learning. The model employs factor-as-piece approximation that divides the linear-chain graph structure into the set of pieces, which preserves the basic Markov Random Field structure and effectively avoids handling large number of candidate output sequences generated by partially annotated data. Then confidence measure is introduced in the model to address different contributions of candidate labels, which enables the ground-truth label information to be utilized in parameter learning. Based on the derived lower bound of the variational lower bound of the proposed model, variational parameters and confidence measures are estimated in the framework of alternating optimization. Moreover, weighted Viterbi algorithm is proposed to incorporate confidence measure to sequence prediction, which considers label ambiguity arose from multiple annotations in the training data and thus helps improve the performance. SGPPSL is evaluated on several sequence labeling tasks and the experimental results show the effectiveness of the proposed model.

\end{abstract}


\IEEEpeerreviewmaketitle

\section{Introduction}
\IEEEPARstart{S}{equence} labeling, which refers to assign a label to each token in a given input sequence, has been successfully applied in Natural Language Processing and Computational Biology. For example, as shown in Figure 1, in Part-Of-Speech (POS) tagging, sequence labeling assigns the POS tag to each word in a given sentence. Figure 1 also demonstrates how sequence labeling detects the mention of genes from biomedical publication abstracts. In these tasks, label assignment for each element should consider the surrounding context of this element in the sequence (e.g. the preceding element and its corresponding label). Sequence labeling can help explore the structure of the given contexts and provide globally optimal label sequence for the input sequence.

\begin{figure}[h]
\centering
\includegraphics[width=3.5in,height = 0.9in ]{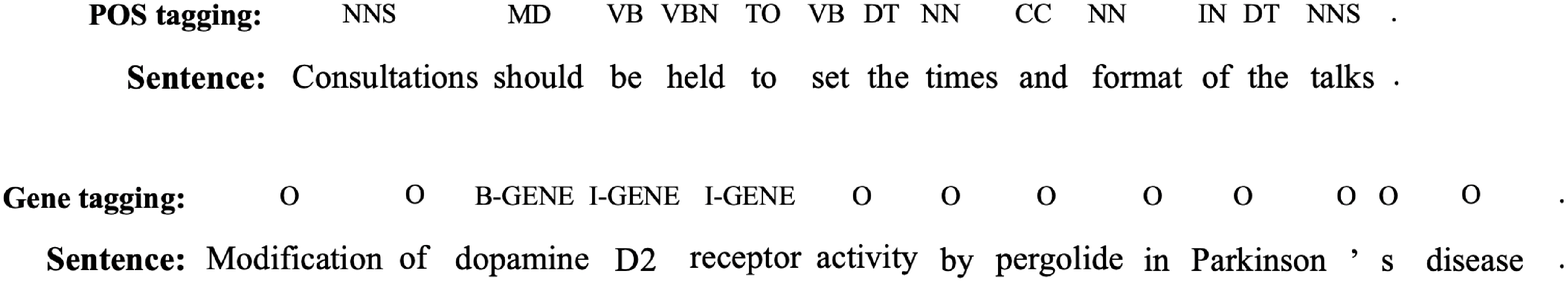}
\caption {Two sequence labeling applications: POS tagging and Gene mention detection, where POS tags include noun (``NNS" and ``NN") and verb (``VB" and ``VBN"), and ``O" stands for outside the Gene entity.}
\end{figure}

Many effective methods like structured support vector machine (S-SVM) and Conditional Random Fields (CRFs) have delivered promising results to sequence labeling. A linear-chain CRFs \cite{re1} directly model the conditional probability of the label sequence without assumption on the dependencies among the observations. They achieve good performance in the tasks of POS tagging and Named Entity Recognition. However, these traditional methods heavily rely on task-specific feature engineering. During the past few years a variety of deep sequence labeling models have been proposed to improve the performance with good feature representation learning. For example, Bi-directional long-short term memory (Bi-LSTM) \cite{re2} is designed to access past features and future features for a specific token in the sequence. Bi-LSTM with CRF on top layer (Bi-LSTM-CRF) \cite{re3} effectively makes use of the label information from past and future tokens to predict the tag of current token. However, the above models usually require large amount of training data with complete annotations \cite{re4,re5,re6}, which is costly and laborious to produce. Although semi-supervised sequence labeling models \cite{re7} can utilize unlabeled data to facilitate learning from a small amount of labeled data, they still need exact annotations.

In reality, it is more cost-effective to obtain a set of candidate labels for the instance \cite{re8}. For example, well-developed commercial crowdsourcing platforms (e.g. Amazon Mechanical Turk (AMT)) provide a cheap and efficient way to obtain large labeled data, where the labeling task is divided into many subtasks that are distributed to a large group of ordinary workers. But the quality of these crowd annotations cannot be guaranteed as the expertise level of workers varies. Figure 2 demonstrates a sentence with crowded named entity labelings on AMT, where each word in the sentence is ambiguously annotated and only annotator 3 provides correct labelings. Therefore utilizing ambiguous annotations to train a prediction model with high performance is practically significant for sequence labeling.

\begin{figure}[h]
\centering
\includegraphics[width=3.2in,height = 0.7in ]{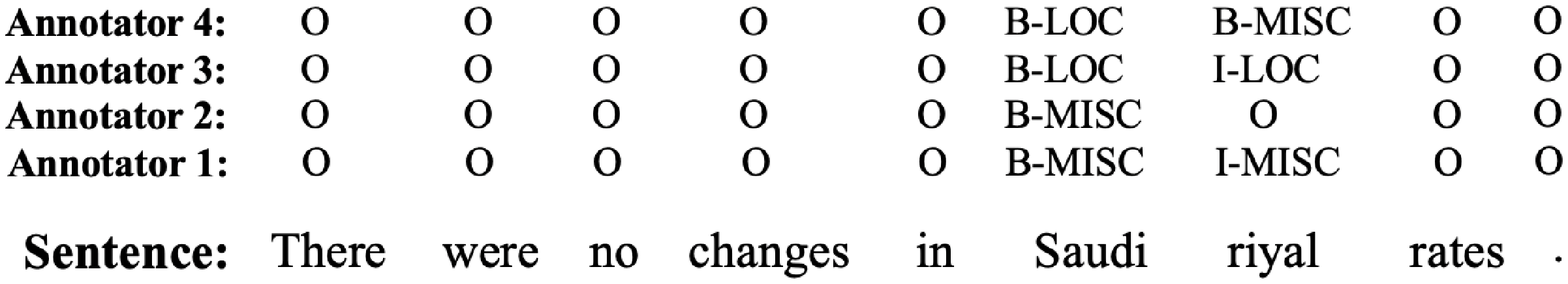}
\caption {A sentence with crowded named entity labelings on AMT, where ``LOC" and ``MISC" refer to location and miscellaneous entity, respectively.} 
\end{figure}

Recently partial label learning (PLL) has been introduced to sequence labeling \cite{re8,re9}. In partial sequence labeling (PSL), the ground-truth label is masked by ambiguous annotations. As shown in Figure 3, each word in a given sentence is annotated with ambiguous POS tags where both the ground-truth and noisy labelings are included. Compared with golden single annotations, the ambiguously annotated sentence is of poor quality as it additionally provides $3^{10}-1$ noisy labelings, which adversely affects the prediction performance of standard supervised learning models. PSL-based methods mainly focus on identifying the ground-truth label from ambiguous annotations. For example, Lou et al. \cite{re9} extended Convex Learning from Partial Labels (CLPL) to sequence labeling that aims to discriminate the ground-truth output sequence from other possible outputs. 

Existing partial sequence labeling models are based on max-margin framework \cite{re8,re9}. While these models are able to deliver good prediction performance by learning from partial annotations, they only produce deterministic outputs and do not quantify the uncertainty in prediction that shows how confident we can be in interpreting the results. In partial sequence labeling, due to inherent model uncertainty \cite{re10} and noisy input,  it is important to measure how confident the PSL model is about prediction. Furthermore, cross validation for model selection and hyperparameter learning in max-margin based PSL models may be computationally infeasible as the search space is too large. For example, there are $n^2$ combinations of two regularization parameters in two large margin formations \cite{re8} for partial sequence labeling, where $n$ is the number of possible values for each regularization parameter.

\begin{figure}[H]
\centering
\includegraphics[width=3.2in,height = 0.65in ]{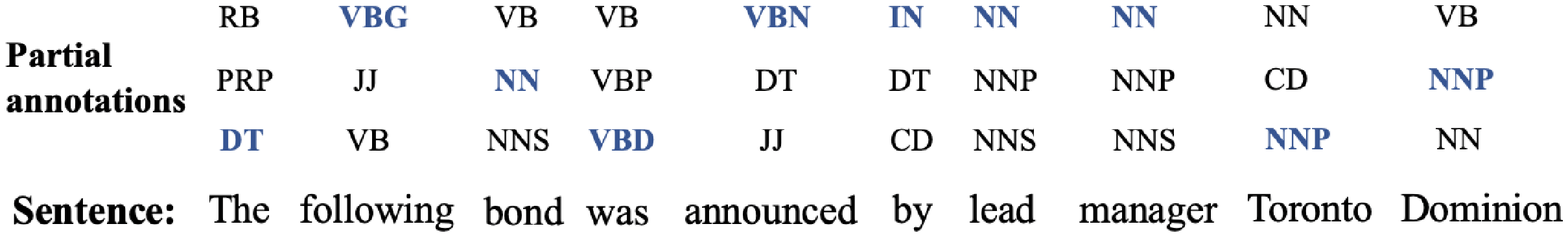}
\caption {An example: POS tagging with partial annotations, where blue tag denotes the ground-truths.} 
\end{figure}

In this paper, to address the above problems in partial sequence labeling, we propose structured Gaussian Processes for partial sequence labeling (SGPPSL) that estimates the posterior distribution of model parameters and posterior predictive distribution, which encodes uncertainty in the prediction and does not need cross validation for model selection and hyperparameter learning. By investigating the combination of Gaussian Processes and CRFs in modeling sequential data, we develop an effective non-parametric Bayesian model to learn from partially annotated sequential data. Our contributions can be summarized as follows:

First, we develop a structured Gaussian Process piecewise-likelihood model. The model employs the factor-as-piece approximation that divides the linear-chain graph structure into the set of pieces, which preserves the basic Markov Random Field structure and effectively avoids handling large number of candidate output sequences generated by partially annotated data. Furthermore, different from unique disambiguation strategy, confidence measure is introduced to address the different contributions of candidate labels, which enables the ground-truth label information to be utilized in parameter learning. 

Second, based on the obtained variational lower bound $L$ of the proposed model, we further derive the lower bound of $L$, which aims to solve the non-differentiable problem in optimization. Variational parameters and confidence measure are estimated in the framework of alternating optimization. 

Third, weighted Viterbi algorithm is proposed to include estimated confidence measures in sequence prediction. For a given test instance, by compressing the confidence measures collected from its nearest neighbors in the training data into a confidence factor and then incorporating it to the score computation in decoding, label ambiguity arose from multiple annotations can be considered in state emission and transition.

Fourth, we evaluate the proposed method on three NLP tasks: Base NP, Chunking and Named Entity Recognition. The experimental results show that in most cases our proposed model outperforms the baselines.

The rest of the paper is organized as follows: Section \Romannum{2} reviews the related work. We describe the formulation, inference and prediction of the proposed structured Gaussian Processes for partial sequence labeling (SGPPSL) in Section \Romannum{3}. A series of experiments are presented in Section \Romannum{4}. Section \Romannum{5} summarizes the paper and discusses the possible future research.

\section{Related work}

Hidden Markov Models (HMMs) \cite{re11,re12} and Conditional Random Fields (CRFs) \cite{re13,re14} are the most widely used graph models for sequence labeling. Many variants of CRFs like semi-Markov conditional random fields (semi-CRFs) \cite{re15} and hierarchical conditional random fields (HCRFs) have been proposed for modeling complex structured outputs. Furthermore, in recently years  deep sequence labeling models that combine deep learning and graph models, such as CRF-CNN \cite{re16} and LSTM-CRF \cite{re17}, have achieved competitive results compared with traditional graph models.

Since supervised based models require large number of training data with exact annotations, semi-supervised sequence labeling has been investigated to effectively utilized large number of unlabeled data for training, which greatly reduces labour costs and improves the efficiency of data collection. For example, semi-supervised CRFs \cite{re7} is proposed to minimize the conditional entropy on unlabeled training instances. Brefeld and Scheffer \cite{re18} developed co-training principle into support vector machine to minimize the number of errors for labeled data and the disagreement for the unlabeled data. While these semi-supervised sequence labeling models partly reduce the demand for large labeled datasets, exact annotations are essential for parameter learning.

Partial label learning (PLL) gets increasing attention to handle one of challenging classification problems where the true label is masked by multiple ambiguous annotations. There has been much research focusing on different disambiguation strategies. The most intuitive way is to treat each candidate label equally and average the scores of all candidate labels. For example, Cour et al. \cite{re19} proposed Convex Loss for Partial Labels (CLPL) to disambiguate candidate labels with non-candidate labels. However, average disambiguation strategy tends to incorporate wrong label information. To identify the true label from candidates, unique disambiguation strategy has been widely accepted. Partial label SVM (PL-SVM) \cite{re20} was formulated to maximize the margin between the best wrong prediction of non-candidate labels and current prediction of the ground-truth label. Yu and Zhang \cite{re21} proposed Maximum Margin Partial Label Learning that aims to address the difference between the ground-truth and other candidates. ``EM+Prior" model \cite{re22} generalizes the EM model with prior knowledge indicating which label is more likely to be the ground-truth and then estimates the ground-truth in iterative optimization. Moreover, Zhang et al. \cite{re23} used Error-Correcting Output Codes (ECOC) coding matrix to treat candidate label set as an entirety (PL-ECOC), which avoids the disambiguation between candidate labels. While PL-ECOC is intuitively simple, the generation of binary training set from partially labeled data heavily relies on coding matrix. Also, nonparametric models have been exploited in partial label learning. Eyke et.al \cite{re24} used K-nearest neighbors weighted voting to predict the ground-truth of the instance. Zhou et.al \cite{re25} applied Gaussian Process to deal with nonlinear classification in partial label learning.

Generalizing partial label learning to sequence labeling, existing work mainly focuses on improving unique disambiguation strategy within the max-margin framework. For example, Lou and Hamprecht \cite{re9} developed CLPL-based model to discriminate candidate label sequences from non-candidate label sequences. Li et al. \cite{re8} proposed CLLP to further discriminate the ground-truth label sequence from other candidate label sequences by maximizing the margin between the ground-truth and other candidates. Although max-margin based models performs well in disambiguation between labels, they cannot provide an uncertainty estimation of the prediction.

In supervised learning, Gaussian Processes (GPs) \cite{re26}, a nonparametric Bayesian model, provides a viable alternative to explicitly capture the uncertainty in prediction with a non-linear Bayesian classifier. Altun et al. \cite{re27} proposed GPs sequence labeling on the basis of GPs multi-class classification, which treats the whole label sequence as an individual label. However, this GPs sequence classification model predicts the new observation sequence with maximum a posterior (MAP) estimation. Further, the large size of possible label set (i.e. the number of possible label sequences) adversely affects the efficiency of the model. Bratieres et al. \cite{re28} proposed GPstruct, a nonparametric structured prediction model. GPstruct combines GPs and CRFs which is used to model the structure imposed by a Markov Random Field. To take advantage of long dependencies in the linear chain structure, Srijith et al. \cite{re29} replaced the likelihood of GPstruct with pesudo-likelihood for sequence labeling (GPSL). 

However, it is infeasible to apply the above GP-based models for partially annotated sequential data. Given a sequence with length $m$ where each token has $l$ candidate labels, these models have to handle $l^m$ candidate output sequences. Furthermore, traditional average and unique disambiguation strategy employed in the nonparametric models \cite{re24,re25} for partial label learning fails to address some candidates that could be false positive or similar to the ground-truth label. As shown in Figure 3, ``following" can be identified as ``JJ" (i.e. adjective) by unique disambiguation, which may adversely affect the learning of prediction model.

\section{Proposed model}

\subsection{Model formulation}
Given sequence data $\left \{ X_{i},Y_{i} \right \}_{i=1}^{N}$, $X^{i}=\left \{ \bm{x}_{1},...,\bm{x}_{m} \right \}$, $Y^{i}=\left \{ \textbf{y}_{1},...,\textbf{y}_{m} \right \}$. $\textbf{y}_{m}=\left\{ y_{1},...,y_{l} \right \}$, where $l$ denotes the number of candidate labels. $\mathcal{Y}$ is the label set for $y$. If $m_{th}$ element in the $i_{th}$ sequence is correctly annotated, then $\textbf{y}_{m}$ is a singleton. 

First, consider the factor graph of linear chain CRF, as shown in Figure 4, where unary factor $\psi (y_{t},\bm{x}_{t})$ and transition factor $\psi (y_{t},y_{t-1},\bm{x}_{t})$ are included. For the partially annotated sequence $X_i$, there are $l^m$ possible candidate output sequences. To avoid handling large number of candidates, we employ piecewise training and use factor-as-piece approximation \cite{re30} to divide the full graph into pieces, where each factor is assigned to its own piece. Then the number of candidate outputs can be reduced to $lm+l^2(m-1)$.

Based on the divided factor graph, the likelihood of $p(Y_{i}|X_{i})$ is defined as 
\begin{equation}
p(Y_{i}|X_{i})=\prod_{a=1}^{M}\prod_{y_{a}^{j}\in S_{a}}C_{a}^{j}p(y_{a}^{j}|\bm{x}_{a}),
\end{equation}
where $M$ is the number of factors in the linear chain structure of $\left \{ X_{i} ,Y_{i}\right \}$. $S_{a}$ denotes the set of candidate labels for $a_{th}$ factor. $C_{a}^{j}$ represents the confidence measure for the $j_{th}$ candidate label in $S_{a}$. The conditional distribution of $p(y_{a}|\bm{x}_{a})$ is defined as 
\begin{equation}
p(y_{a}|\bm{x}_{a})=
\left\{
\begin{array}{lcl}
\frac{\psi(y_a,\bm{x}_a)}{\sum\nolimits_{y'}\psi(y',\bm{x}_a) }, \mathrm{unary \ factor},\\

\frac{\psi(y_a,y_{a-1},\bm{x}_a)}{\sum\nolimits_{y'}\psi(y',y_{a-1},\bm{x}_a)}, \mathrm{transition \ factor}.

\end{array} \right.
\end{equation}

\begin{figure}[H]
\centering
\includegraphics[width=2.8in,height = 1in ]{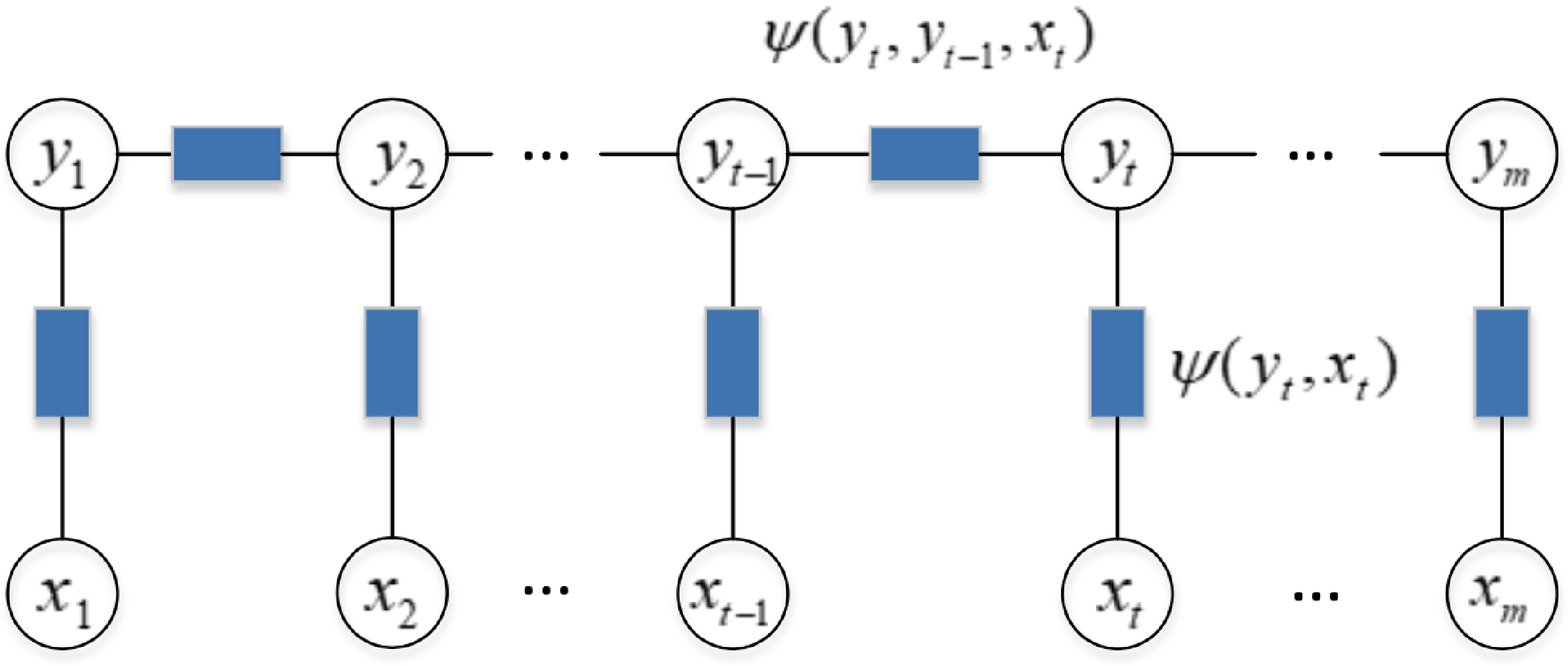}
\caption {An example: POS tagging with partial annotations.} 
\end{figure}

As shown in Equation (1), the modeling structured output for each factor is weighted by $C_{a}^{j}$, which addresses the different contributions of candidate labels. Compared with unique ground-truth identification strategy, confidence weighted mechanism enables the ground-truth label to be utilized in the learning process.

GP classification introduces latent variables (LVs) to mediate the influence of the input and the output. We define LV per factor. There are two types of latent variables corresponding to different factor types in the linear-chain structure: the unary LV $f_u$ and the transition LV $f_t$. $\mathbf{f}_{u}\left ( y \right )=\left \{ f_{u}(\bm{x}_1,y) ,...,f_{u}(\bm{x}_{N\times m\times l},y)\right \}$ where $f_{u}\left (\bm{x},y \right )$ is associated with a specific label $y$ for each training instance $\bm{x}$. The transition LV $f_t$ is defined over all label pairs $\left ( y,y' \right )$ where $y,y'\in \mathcal{Y}$. Based on the defined latent variables, given partially annotated sequence data $\left \{ X_{i},Y_{i} \right \}_{i=1}^{N}$, the likelihood of $p(Y|X,\textbf{f})$ is defined as 
\begin{equation}
p(Y|X,\textbf{f})=\prod_{i=1}^{N}\prod_{a=1}^{M_i}\prod_{y_{a}^{j}\in S_{ia}}C_{ia}^{j}\frac{\exp(f(\bm{x}_{ia},y_{ia}^{j}))}{\sum_{y^{'}}\exp(f(\bm{x}_{ia},y^{'}))},
\end{equation}
where $\mathbf{f}=\left \{ \mathbf{f}_U,\mathbf{f}_T \right \}$. $\mathbf{f}_T$ is the collection of $f_t$. The number of $f_t$ is fixed with $\left | \mathcal{Y} \right |^{2}$ while the number of $f_u$ grows with the size of data. $\mathbf{f}_U$ is the collection of $\mathbf{f}_{u}$ defined over all possible labels in $ \mathcal{Y}$. There are total $N\times m\times l\times \left | \mathcal{Y} \right |$ unary LVs in $\mathbf{f}_U$.

\subsection{GP prior and posterior}
The latent variables $\mathbf{f}_{u}\left ( y \right )$ is drawn from a zero mean GP with covariance function $K_u\left ( y \right )$ of size $Nml\times Nml$ where $K_{u(i,j)}(y)= k((\bm{x}_i,\bm{x}_j);\theta _y)$. $k((\bm{x}_i,\bm{x}_j);\theta _y)$ is defined as
\begin{equation}
k((\bm{x}_i,\bm{x}_j);\theta _y)=  \exp(-\theta_y \left \| \bm{x}_i-\bm{x}_j\right \|^{2}).
\end{equation}
where $\theta _y$ is a kernel hyperparameter.

The transition latent variables $\mathbf{f}_T$ is defined with a zero mean and covariance $K_T$. $K_T(i,j)$ is a covariance function of label pairs $(y_i,y'_i)$ and $(y_j,y'_j)$,which takes the following form
\begin{equation}
\llbracket (y_{i}=y_{j}\wedge y'_{i}=y'_{j}) \rrbracket .
\end{equation}

Based on the above specification for $\mathbf{f}_U$ and $\mathbf{f}_T$, the GP prior over \textbf{f} is defined as
\begin{equation}
 \begin{split}
p(\textbf{f}|X) &=  N\left (\textbf{f};0,K \right)\\
&= N\left ( \begin{bmatrix} \textbf{f}_{U}\\ \textbf{f}_{T}\end{bmatrix} ;0,\begin{bmatrix}K_{U} &0 \\ 0& K_{T}\end{bmatrix}\right ),         
 \end{split}
\end{equation}
where $K_U$ is block-diagonal with $\left | \mathcal{Y} \right |$ equal size blocks. $K_T = I_{\left | \mathcal{Y} \right |^{2}}$.

The posterior distribution over \textbf{f} is 
\begin{equation}
p(\textbf{f}|X,Y)=\frac{p(Y|X,\textbf{f})p(\textbf{f}|X)}{p(Y|X)},
\end{equation}
where 
\begin{equation}
p(Y|X)=\int p(Y|X,\textbf{f})p(\textbf{f}|X)d\textbf{f}.
\end{equation}

\subsection{Variational Gaussian approximate inference}

The posterior distribution $p(\textbf{f}|X,Y)$ cannot be calculated analytically as the non-Gaussian property of $p(Y|X,\textbf{f})$. The most common way is to approximate the posterior with a tractable Gaussian distribution. 

First, variational inference approximates $p(\textbf{f}|X,Y)$ with a variational distribution $q(\textbf{f})$ by using the following criterion:
\begin{equation}
\min\limits_{q(\textbf{f})} KL(q(\textbf{f})||p(\textbf{f}|X,Y)),
\end{equation}
where $KL$ refers to Kullback-Leibler (KL) divergence.

The log-likelihood of $p(Y|X)$ can be derived as follows:

\begin{equation}
 \begin{split}
 \log p(Y|X) &= \int q(\textbf{f})\log p(Y|X)d\textbf{f}\\
&= \int q(\textbf{f}) \log \frac{p(Y|X,\textbf{f})p(\textbf{f}|X)q(\textbf{f})}{p(\textbf{f}|X,Y)q(\textbf{f})}d\textbf{f}\\
&=  \scalebox{0.9}{$ \int q(\textbf{f}) \log \frac{q(\textbf{f})}{p(\textbf{f}|X,Y)}d\textbf{f} + \int q(\textbf{f}) \log \frac{p(Y|X,\textbf{f})p(\textbf{f}|X)}{q(\textbf{f})}d\textbf{f} $}\\
&= KL(q(\textbf{f})||p(\textbf{f}|X,Y)) + L(q(\textbf{f})),
 \end{split} 
\end{equation}
where $L(q(\textbf{f}))$ is evidence lower bound (ELBO).

We have $\log p(Y|X) \geq L(q(\textbf{f}))$ due to the non-negative property of $ KL(q(\textbf{f})||p(\textbf{f}|X,Y))$. Then the final optimization of variational inference is to maximize $L(q(\textbf{f}))$ which can be defined as

\begin{equation}
 \begin{split}
 L(q(\textbf{f})) &= \int q(\textbf{f}) \log \frac{p(\textbf{f}|X )p(Y|X,\textbf{f})}{q(\textbf{f})}d\textbf{f}\\
                       &= -KL(q(\textbf{f})||p(\textbf{f}|X))+\int q(\textbf{f}) \log p(Y|X,\textbf{f})d\textbf{f}\\
                       &= -KL(q(\textbf{f})||p(\textbf{f}|X)) \\
                       &+ \sum_{i=1}^{N}\sum_{a=1}^{M_i}\sum_{y_{a}^{j}\in S_{ia}} \left [  \mathbb{E}_{q(\textbf{f})}\left [ \log p(y_{ia}^{j}|\bm{x}_{ia},\textbf{f}) \right ] + \log (C_{ia}^{j}) \right ].
 \end{split}
\end{equation}

Variational Gaussian approximate inference \cite{re31} assumes the posterior $q(\textbf{f})$ to be a multivariate Gaussian:
\begin{equation}
 \begin{split}
q(\textbf{f}) &=  N\left (\textbf{f};\mu,V \right)\\
&= N\left ( \begin{bmatrix} \textbf{f}_{U}\\ \textbf{f}_{T}\end{bmatrix} ;\begin{bmatrix} \mu_{U}\\ \mu_{T}\end{bmatrix},\begin{bmatrix}V_{U} &0 \\ 0& V_{T}\end{bmatrix}\right ).     
 \end{split}
\end{equation}

Based on the closed formed expression of the KL divergence between two Gaussians, $KL(q(\textbf{f})||p(\textbf{f}|X))$ can be written as
\begin{equation}
\scalebox{0.92}{$
KL(q(\textbf{f})||p(\textbf{f}|X)) = \frac{1}{2}\left [ \log\left |V^{-1}K  \right |  + tr(VK^{-1})-d+\mu^{T}K^{-1}\mu\right]$},
\end{equation}
where $d$ is a constant and equals the number of variational parameters.

The expectation $\mathbb{E}_{q(\textbf{f})}\left [ \log p(y_{a}^{j}|\bm{x}_{a},\textbf{f}) \right ]$ is intractable as $p(y_{a}^{j}|\bm{x}_{a},\textbf{f})$ is a softmax function. We employ Jensen's inequality to obtain the tractable lower bound of $\mathbb{E}_{q(\textbf{f})}\left [ \log p(y_{a}^{j}|\bm{x}_{a},\textbf{f}) \right ]$ which is stated as follows:
\begin{equation}
 \begin{split}
 &\mathbb{E}_{q(\textbf{f})}\left [ \log p(y_{a}^{j}|\bm{x}_{a},\textbf{f}) \right ] =\mathbb{E}_{q(\textbf{f})}\left [ \log \frac{\exp(f(\bm{x}_{a},y_{a}^{j}))}{\sum\nolimits_{y'}\exp(f(\bm{x}_{a},y'))} \right ] \\
&=\mathbb{E}_{q(\textbf{f})}\left [ f(\bm{x}_{a},y_{a}^{j}) \right ]-\mathbb{E}_{q(\textbf{f})}\left [ \log \sum\nolimits_{y'}\exp(f(\bm{x}_{a},y'))\right ]\\
&\geq  \mathbb{E}_{q(\textbf{f})}\left [ f(\bm{x}_{a},y_{a}^{j}) \right ]-\log \sum\nolimits_{y'}\mathbb{E}_{q(\textbf{f})}\left [  \exp(f(\bm{x}_{a},y')\right ]\\
&= \mu(\bm{x}_{a},y_{a}^{j})-\log \sum\nolimits_{y'}\exp\left [ \mu(\bm{x}_{a}^{j},y') +\frac{1}{2}V_{\left ( (\bm{x}_{a}^{j},y'),(\bm{x}_{a}^{j},y')\right )}\right ],
  \end{split}
\end{equation}
where $\bm{x}_{a}^{j}$ refers to the $a_{th}$ factor tagged with $j_{th}$ candidate label.

Then the optimization problem turns into maximizing the lower bound of $L(q(\textbf{f}))$ which is defined as
\begin{equation}
\begin{split}
L_{l}(q(\textbf{f}))&=\frac{1}{2}\left [ \log\left |VK^{-1}  \right |  - tr(VK^{-1})+d-\mu^{T}K^{-1}\mu\right ]\\
&+ \sum_{i=1}^{N}\sum_{a=1}^{M_i}\sum_{y_{a}^{j}\in S_{ia}} \mu(\bm{x}_{ia},y_{ia}^{j})\\
&-\log \sum\nolimits_{y'}\exp\left [ \mu(\bm{x}_{ia}^{j},y') +\frac{1}{2}V_{\left ( (\bm{x}_{ia}^{j},y'),(\bm{x}_{ia}^{j},y')\right )}\right] \\
&+ \sum_{i=1}^{N}\sum_{a=1}^{M_i}\sum_{y_{a}^{j}\in S_{ia}} \log (C_{ia}^{j}),
\end{split}
\end{equation}
where for the formula taking the form $\mu(\bm{x}_t,y_t)-\log \sum\nolimits_{y'}\exp\left [ \mu(\bm{x}_t,y') +\frac{1}{2}V_{\left ( (\bm{x}_t,y'),(\bm{x}_t,y')\right )}\right]$ is computed with
\begin{equation}
\begin{split}
&\textbf{unary \ factor}: \\
&\mu_{\scalebox{0.5}{$U_t$}}(y)-\log \sum\nolimits_{y'}\exp\left [ \mu_{\scalebox{0.5}{$U_t$}}(y') +\frac{1}{2}V_{U_{(t,t)}}(y')\right], \\ 
&\textbf{transition \ factor}:\\
&\scalebox{0.85}{$\mu_{\scalebox{0.5}{$T$}}(y_{t-1},y_t)-\log \sum\nolimits_{y'}\exp\left [ \mu_{\scalebox{0.5}{$T$}}(y_{t-1},y') +\frac{1}{2}V_T{((y_{t-1},y'),(y_{t-1},y'))}\right] $}.
\end{split}
\end{equation}

\subsection{Parameter Estimation}
The parameters $\left \{ (\mu,V),\bm{\theta},C \right \}$ are required to be estimated, where $\bm{\theta}$ is the set of $\left \{ \theta _y \right \}$ for $K$ and $C=\left \{ c_{a}^{j} \right \}$ denotes the set of confidence measure of each candidate labels for a training instance. Alternating optimization is employed to estimate parameters:

1. Fixed $\left \{ (\mu,V),\bm{\theta } \right \}$, each element $ c_{a}^{j}$ in $C$ is computed with
\begin{equation}
c_{a}^{j}= \frac{\exp(\mu(\bm{x}_{a},y_{a}^{j}))+\frac{1}{2}V_{((\bm{x}_{a},y_{a}^{j}),(\bm{x}_{a},y_{a}^{j}))}}{\sum_{j}\exp(\mu(\bm{x}_{a},y_{a}^{j}))+\frac{1}{2}V_{((\bm{x}_{a},y_{a}^{j}),(\bm{x}_{a},y_{a}^{j}))}}.
\end{equation}

$p(y_a|\bm{x}_a)$ takes the form of $\int p(y_a|\bm{x}_a,\mathbf{f})p(\mathbf{f})d\mathbf{f}$, where the computation of expectation of softmax function is intractable. We employ the strategy \cite{re29} that computing softmax of the expectation for latent variables to obtain the probability of $p(y_a|\bm{x}_a)$, and then compute $c_{a}^{j}$ as shown in Equation (17).

2. Fixed $C$. Optimizing $\left \{ (\mu,V),\bm{\theta } \right \}$ by solving the maximization of $L_{l}(q(\textbf{f}))$.
Based on the concavity of  $L_{l}(q(\textbf{f}))$, parameter optimization can be realized by the nested loop. In the inner loop, we use co-ordinate ascent algorithm to estimate $\left \{ (\mu,V) \right \}$ with fixed $\bm{\theta }$. Then the hyperparamter $\bm{\theta }$ is estimated in the outer loop for the fixed $\left \{ (\mu,V) \right \}$. The gradients w.r.t $\left \{ (\mu,V),\bm{\theta } \right \}$ is computed as follows:
\begin{equation}
\triangledown_{\mu}L_l=-K^{-1}\mu+\frac{\partial P}{\partial \mu_{(t,iaj)}},
\end{equation}
\begin{equation}
\triangledown_{V}L_l=\frac{1}{2}\left ( V^{-1}-K^{-1} \right )+\frac{\partial P}{\partial V_{(t,iaj)}},
\end{equation}
\begin{equation}
\scalebox{0.9}{$
\triangledown_{\bm{\theta}_t}L_l=\left ( K^{-1}\mu \right )\frac{\partial K}{\partial \bm{\theta}_t}\left ( K^{-1}\mu \right )^{T}+tr\left [ K^{-1} \left ( VK^{-1}-I \right )\frac{\partial K}{\partial \bm{\theta}_t}\right ]$},
\end{equation}
where $t$ refers $t_{th}$ block as there are total $\left | \mathcal{Y} \right | + 1$ blocks in $V$, and 

\begin{equation}
\begin{split}
P = &\sum_{i=1}^{N}\sum_{a=1}^{M_i}\sum_{y_{a}^{j}\in S_{ia}} \mu_(\bm{x}_{ia},y_{ia}^{j})\\
&-\log \sum\nolimits_{y'}\exp\left [ \mu(\bm{x}_{ia}^{j},y') +\frac{1}{2}V_{\left ( (\bm{x}_{ia}^{j},y'),(\bm{x}_{ia}^{j},y')\right )}\right].\\
\end{split}
\end{equation}

\subsection{Prediction}

Given the estimated parameters $\left \{ (\mu,V),\bm{\theta },C \right \}$ and a new sequence data $\left \{ \mathbf{x}_{*},\mathbf{y}_{*} \right \}$, the posterior distribution $p(\mathbf{f}_*|X,Y,\mathbf{x}_{*})$ is defined as 
\begin{equation}
\begin{split}
p(\mathbf{f}_*|X,Y,\mathbf{x}_{*})&=\int p(\mathbf{f}_*|X,\mathbf{x}_{*},\mathbf{f})p(\mathbf{f}|X,Y)d\mathbf{f}\\
&=N\left ( \begin{bmatrix} \textbf{f}_{*U}\\ \textbf{f}_{*T}\end{bmatrix} ;\begin{bmatrix} \mu_{*U}\\ \mu_{*T}\end{bmatrix},\begin{bmatrix}V_{*U} &0 \\ 0& V_{*T}\end{bmatrix}\right ),\\
\end{split}
\end{equation}
where $\mu_{*T}=\mu_{T}$ and $V_{*T} = V_{T}$. $\mu_{*U}$ and $V_{*U}$ are derived as
\begin{equation}
\mu_{*U}=K_{*U}K_{U}^{-1}\mu_{U},
\end{equation}
\begin{equation}
 V_{*U}=K_{*U}-K_{*U}^T(K_U^{-1}-K_U^{-1}V_{U}K_U^{-1})K_{*U}.
\end{equation}
The predictive probability of $p(\mathbf{y}_{*i}|\mathbf{x}_{*i},X,Y)$ is defined as
\begin{equation}
p(\mathbf{y}_{*i}|\mathbf{x}_{*i},X,Y)=\int p(\mathbf{y}_{*i}|\mathbf{f}_{*U}) p(\mathbf{f}_{*U}|X,Y,\mathbf{x}_{*})d\mathbf{f}_{*U},
\end{equation}
where $p(\mathbf{y}_{*i}|\mathbf{f}_{*U})$ is defined as
\begin{equation}
p(\mathbf{y}_{*i}|\mathbf{f}_{*U}) = \frac{\exp(f_{*Ui}(\mathbf{y}_{*i}))}{\sum\nolimits_{y'}\exp(f_{*Ui}(y'))}.
\end{equation}
Due to the intractability of the expectation of softmax function, we also use the same strategy stated in Equation (17) to define the score for assigning $\mathbf{y}_{*i}$ to $\mathbf{x}_{*i}$ and transiting from $\mathbf{y}_{*(i-1)}$ to $\mathbf{y}_{*i}$ as 
\begin{equation}
S(\mathbf{y}_{*i},\mathbf{x}_{*i})=\frac{\exp(\mu_{*Ui}(\mathbf{y}_{*i})+\frac{1}{2}V_{*Ui}(\mathbf{y}_{*i}))}{\sum_{y'}\exp(\mu_{*Ui}(y')+\frac{1}{2}V_{*Ui}(y'))},
\end{equation}
\begin{equation}
\scalebox{0.98}{$
S(\mathbf{y}_{*(i-1)},\mathbf{y}_{*i})=\frac{\exp(\mu_{*T}(\mathbf{y}_{*(i-1)},\mathbf{y}_{*i})+\frac{1}{2}V_{*T}(\mathbf{y}_{*(i-1)},\mathbf{y}_{*i}))}{\sum\nolimits_{y'}\exp(\mu_{*T}(\mathbf{y}_{*(i-1)},y')+\frac{1}{2}V_{*T}(\mathbf{y}_{*(i-1)},y'))}$}.
\end{equation}

Based on the confidence measures collected from the K-nearest neighbors of $\mathbf{x}_{*i}$ in the training data, we define the confidence factor $\tau _i(y_*)$ that denotes the confidence of assigning the label $y_*$ to $\mathbf{x}_{*i}$ as
\begin{equation}
\tau _i(y_*)=
\left\{
\begin{array}{lcl}
\frac{1}{K_i}\sum_{k=1}^{K_i}c_{k}(y_*),\ y_*\in\mathcal{Y}_i ,\\
\frac{1}{K_i}\sum_{k=1}^{K_i}\sum \nolimits_{y}c_{k}(y),\ y_*\in\mathcal{Y}\setminus \mathcal{Y}_i,\\
\end{array} \right.
\end{equation}
where $\mathcal{Y}_i $ denotes the set of labels collected from the K-nearest neighbors of $\mathbf{x}_{*i}$ in the training data. Since $\left | \mathcal{Y}_i \right |\leq \left | \mathcal{Y} \right |$, for the assigned label that is not included in $\mathcal{Y}_i$, we choose to average the summarization of all confidence measures obtained from nearest neighbors, which guarantees that the confidences of these labels are lower than that of the ground-truth label.

Then the confidence factor of label transition $\tau (y_1,y_2)$ is defined as $\frac{1}{T}\sum_{t=1}c(y_1,y_2)$, where $T$ denotes the number of transition factor taking the form $(y_1,y_2)$ in the training data. In most cases partially annotated data covers diverse labels, for the label transition that is not included in the training data, we set the corresponding confidence measure to 0 to avoid invalid label transition.

For the weighted Viterbi decoding, the score $g_i(y_{(i-1)},y_i)$ is defined as
\begin{equation}
\scalebox{0.9}{$
g_i(y_{(i-1)},y_i)=\tau _i(\mathbf{y}_{*i})S(\mathbf{y}_{*i},\mathbf{x}_{*i})+\tau (\mathbf{y}_{*(i-1)},\mathbf{y}_{*i})S(\mathbf{y}_{*(i-1)},\mathbf{y}_{*i})$}.
\end{equation}

In Viterbi recursion, the optimal intermediate score $\delta _{t}(s)$ for $t_{th}$ token with label $s$ is represented with
\begin{equation}
\delta _{t}(s)=\max\limits_{y_{t-1}\in \mathcal{Y}}\left [ \delta _{t-1}(y_{t-1}) + g_t(y_{t-1},s)\right ].
\end{equation}

Finally the optimal label sequence can be obtained by path backtracking.

\section{Experimental Results}
In this section, we perform experiments on three NLP tasks: Base NP, Chunking, and Named Entity Recognition. These tasks aim to find the meaningful segments from input sequences, which greatly benefits most of NLP applications, such as document summarization and question answering. However, in these tasks exact annotations for collecting training data are not feasible because of words' ambiguity. For example, annotators may be confused by ``sprout" in identifying noun or verb phrase. Thus applying partial sequence labeling can effectively handle ambiguous label annotations.

\subsection{Datasets}

Base NP: This task is to identify noun phrases for a given sequence, which can be used for many downstream tasks. For example, keyphrase extraction can be treated as assigning the identified noun phrase with the label ``keyphrase" or ``non-keyphrase". We use the dataset developed by Lance Ramshaw and Mitch Marcus \cite{re32} in the NP chunking experiments.

Chunking: This task divides the sentence into smaller segments which include noun phrase, verb phrase and prepositional phrase.  We choose the chunking dataset from CoNLL- 2000 shared task \cite{re33}. This dataset is collected from Wall Street Journal (WSJ) sections, where sections 15-18 as training data (211727 tokens) and section 20 as test data (47377 tokens). 

Named Entity Recognition (NER): This task is to identify the entity types in the sentence. We choose the NER dataset from CoNLL-2003 shared task \cite{re34}. This dataset contains four entity types:  location, organization, person and miscellaneous.

Table \Romannum{1} summarizes the details of three datasets. We further construct a relative small size dataset for each task which aims to avoid large matrix operation for sequential data. Further, the setting of partial annotations is included, where $cl$ refers to the number of candidate labels and $p$ is the proportion of exactly annotated instances. For the generation of candidate labels, as there is no prior knowledge about the ambiguous annotations toward the target label, we conduct random sampling $cl$ times and each randomly select a negative label to the candidates.

\renewcommand{\arraystretch}{1.5}
\begin{table}
\centering
\caption{Summarization of three datasets.}
\begin{tabular}{l|c|c|l} 
\hline
Task     & \#Sentences~ & \#Labels & \multicolumn{1}{c}{Setting of partial annotation}                                                                               \\ 
\hline
Base NP  & ~500         & 3        & \multirow{3}{*}{\begin{tabular}[c]{@{}l@{}}$cl=\left \{ 2,3,4 \right \}$\\$p=\left \{ 0.1,0.3,...,0.9 \right \}$\end{tabular}}  \\
Chunking & ~450         & 20       &                                                                                                                                 \\
NER      & ~500         & 9        &                                                                                                                                 \\
\hline
\end{tabular}
\end{table}

\iffalse
\begin{table}
\centering
\caption{Summarization of three datasets.}
\begin{tabular}{l|c|c|l} 
\hline
\multicolumn{1}{c|}{Task} & \#Sentences~ & \#Labels                 & \multicolumn{1}{c}{Setting of partial annotation}                                                                                                       \\ 
\hline
Base NP     & ~500        & 3                        & \multirow{3}{*}{\begin{tabular}[c]{@{}l@{}} $cl=\left \{ 2,3,4 \right \}$\\$p=\left \{ 0.1,0.3,...,0.9 \right \}$ \end{tabular}}  \\
Chunking                  & ~450         & 20 &                                                                                                                                   \\
NER                       & ~500       & 9                        &                                                                                                                                   \\
\hline
\end{tabular}
\end{table}
\fi

\subsection{Baselines}

We compare the proposed SGPPSL with the following baselines. These baselines include parametric and non-parametric models. Also, the disambiguation strategy employed in these models can be classified into four types: random disambiguation, average disambiguation (AD), identification disambiguation (ID) and disambiguation-free. Representative approaches are stated as follows:

NAIVE \cite{re8}: structured SVM model that randomly chooses a label from candidates ad the ground-truth label.

CLPL \cite{re9}: AD-based method that maximizes the margin between candidate labels and non-candidate labels.

K-nearest weighted voting (WKNN) \cite{re24}: a non-parametric approach which assigns the nearest neighbors with different weight based on distance and treats candidate labels equally (AD type). [suggested setup: k = 5].

PL-ECOC \cite{re23}: a disambiguation-free strategy by encoding partially labeled instances with error-correcting output codes (ECOC) [suggested setup: the codeword length $L =\left \lceil 10\cdot \log_{2} (q)\right \rceil$.

PL-SVM \cite{re20}: ID-based method that maximizes the margin between the ground-truth label and the best prediction of wrong label.

CLLP \cite{re8}: ID-based method which incorporates two types of margins: the margin between the ground-truth label and other candidate labels; the margin between the ground-truth label and non-candidate labels.

PALOC \cite{re35}: ID-based method which induces the multi-class classifiers with one-vs-one decomposition strategy by considering the relevancy of each label pair in the candidate label set.

SURE \cite{re36}: a self-training based unified framework that utilizes the maximum infinity norm regularization to jointly train the predictive model and perform pseudo-labeling (ID type). 

PL-AGGD \cite{re37}: feature-aware disambiguation that jointly identifies the ground-truth label, determines similarity graph and learns predictive model (ID type).

To measure the generalization performance of the baselines and SGPPSL, we employ 5-fold cross validation strategy to train these models on the selected datasets and report the average performance. Specifically, for SGPPSL cross validation is set as an outer loop of the framework of parameter estimation as described in Section \Romannum{3} D.

\subsection{Recovering the ground-truth label}

The estimated confidence measure $C$ can be used to determine the ground-truth label for each token in the sequence, where $y_{i}=\arg\max\limits_{y_j\in S_a}c_{a}^{j}$. Table \Romannum{2} records the average accuracy of SGPPSL on recovering the ground-truth label for training data. With the increase of proportion of annotated instances, in most cases the accuracy of recovering the ground-truth label is improved. Further, the performance is negatively corrected with the number of candidate labels. Multiple annotations based on diverse backgrounds increase label ambiguity.

To demonstrate the confidence evolution in alternating optimization, we randomly select a group from 5-fold cross validation configurations. We record the change of confidence values of candidate labels for some training instances under several special settings: in Base NP ``holding" from the sentence ``...is little holding sterling firm at.." with $\left \{[``B",``O", ``\textbf{I}"],p=0.1 \right \}$, ``German" from the sentence ``...it disagreed with German advice..." with $\left \{[``B-LOC", ``\textbf{B-MISC}", ``B-ORG"], p=0.5 \right \}$ for NER, and in Chunking ``heavily" from the sentence ``... figure are very heavily on..." with $\left \{[``\textbf{I-ADVP}", ``B-ADVP", ``B-PRT"], p = 0.5 \right \}$, where the ground-truth label is highlighted in bold.

\iffalse
In Base NP we choose ``holding" with candidate labels $[``I",``O"]$ and ``sale" with $[``B",``O",``I",``O"]$. ``expected" with candidate labels $[``B-VP", ``I-VP"]$ and ``heavily" with $[``I-ADVP", ``B-ADVP", ``B-PP", ``B-PRT"]$ are select in Chunking task. For NER ``German" with $[``B-LOC", ``B-MISC"]$ and ``EU-wide" with $[``B-LOC", ``B-ORG", ``B-MISC", ``I-ORG"]$ are selected. 
\fi

\begin{figure}[h]
\centering
\includegraphics[width=3.7in,height = 1.3in ]{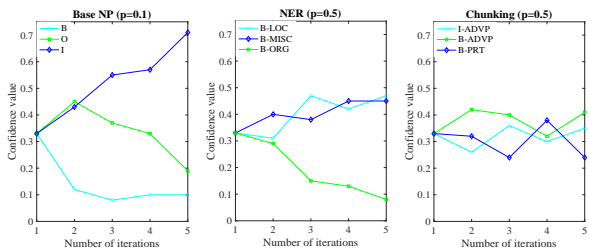}
\caption {Examples of confidence evolution in alternating optimization.}
\label{fig:secondfigure}
\end{figure}

As shown in Figure 5, we can see that ``holding" can be correctly assigned with ``I" even with little exactly annotated training samples in Base NP while ``German" and ``heavily" are misclassified with half exact annotations. Compared with relative large label set, limited label space (e.g. Base NP task) does not add too much noise in partial annotations. This helps reduce confusing label information for disambiguation. Furthermore, existing partial sequence labeling models employ unique identification strategy, which may select wrong ground-truth label for parameter learning. The proposed SGPPSL introduces confidence measure to address different contributions of candidate labels, which enables the ground-truth label to be utilized in the learning process.

\renewcommand{\arraystretch}{1.5}
\begin{table}
\centering
\caption{The performance of recovering the ground-truth label (\%).}
\begin{tabular}{lcccccc} 
\hline
\multicolumn{2}{l}{Task}           & 0.1   & 0.3   & 0.5   & 0.7   & 0.9    \\ 
\hline
\multirow{3}{*}{Base NP}  & $cl=2$ & 72.04 & 76.68 & 84.06 & 86.71 & 89.13  \\
                          & $cl=3$ & 73.35 & 76.63 & 79.02 & 80.39 & 90.83  \\
                          & $cl=4$ & 75.30 & 72.67 & 75.56 & 78.73 & 84.17  \\ 
\hline
\multirow{3}{*}{Chunking} & $cl=2$ & 70.85 & 72.07 & 72.42 & 73.94 & 75.79  \\
                          & $cl=3$ & 62.27 & 64.70 & 67.13 & 67.54 & 70.20  \\
                          & $cl=4$ & 60.50 & 60.62 & 62.87 & 65.10 & 65.97  \\ 
\hline
\multirow{3}{*}{NER}      & $cl=2$ & 73.35 & 73.77 & 74.69 & 84.43 & 87.23  \\
                          & $cl=3$ & 70.88 & 72.38 & 72.06 & 73.75 & 77.61  \\
                          & $cl=4$ & 69.30 & 69.05 & 70.39 & 73.77 & 73.86  \\
\hline
\end{tabular}
\end{table}

\subsection{Comparing with the baselines}

By varying $cl$ from 2 to 4 and $p$ from 0.1 to 0.9, we report the average F1 score that has been widely used in NLP tasks \cite{re15} to measure prediction performance on test data.

1) Base NP.

Figure 6 presents the performance of compared methods on the Base NP task. We can observe that SGPPSL outperforms the most of baselines. PL-SVM, PL-AGGD and SURE achieve competitive results with lower $p$ (e.g. $p = 0.1$). While the average performance of $p > 0.5$ is always better than that of $p < 0.5$, there is no significant positive (or negative) relationship between the performance and the proportion of exactly annotated training instances for most of partial label learning based methods. Furthermore, the performance of PSL based methods does not vary significantly as the number of candidate labels increases.

It is worth noting that the performance of most of compared methods are very close. For example, SGPPSL outperforms the baselines in $cl = 2$ setting by a small margin. As the size of label set for Base NP task is very small (i.e. [``B", ``I", ``O"]), there is not much confusing label information in partial annotations, which may decrease disambiguating difficulty for PSL models. Thus most of the compared methods can achieve good performance even with lower $p$.

\begin{figure*}[!htb]
\minipage{0.32\textwidth}
  \includegraphics[width=\linewidth]{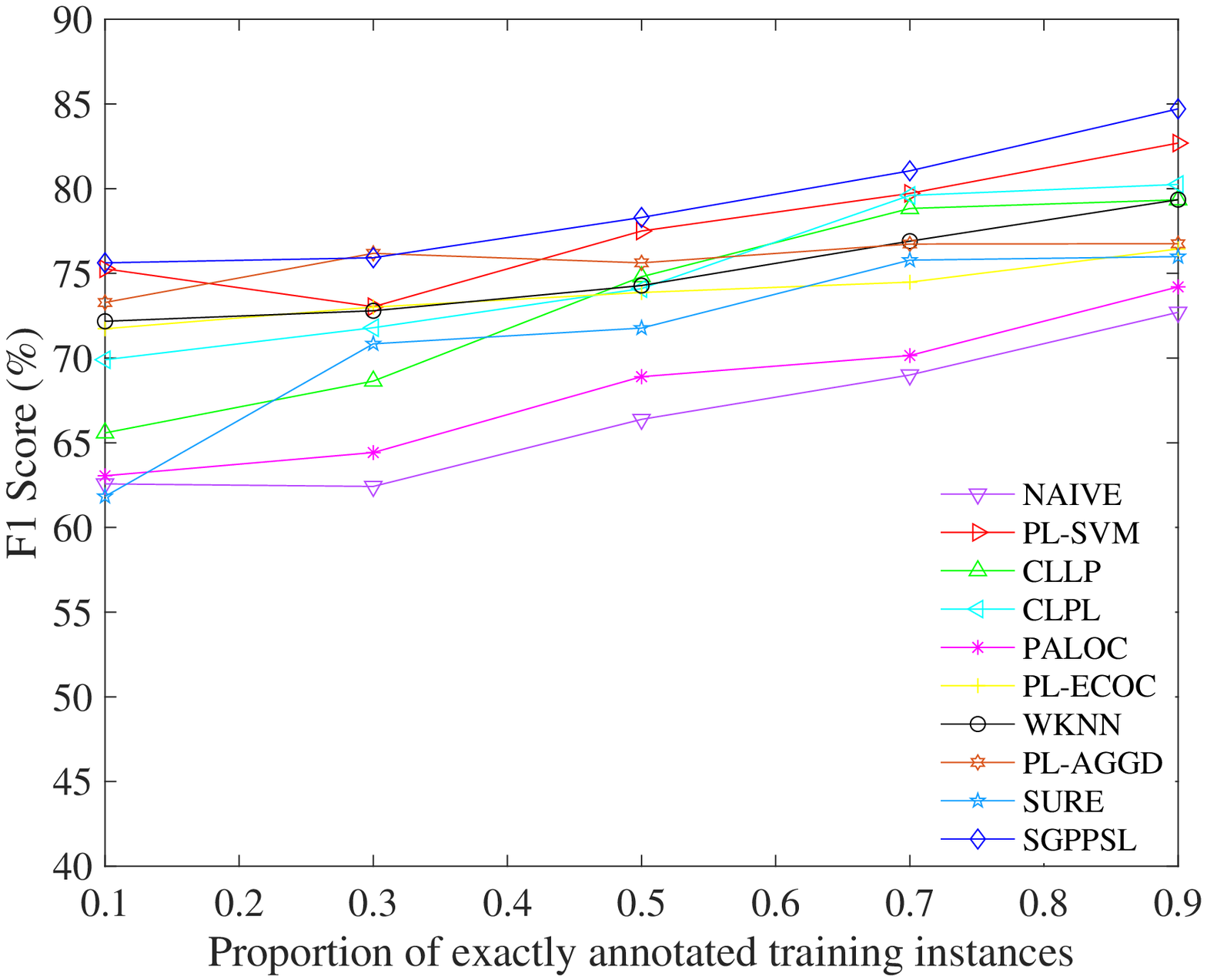}
  \subcaption{$cl$ = 2}\label{fig:awesome_image1}
\endminipage\hfill
\minipage{0.32\textwidth}
  \includegraphics[width=\linewidth]{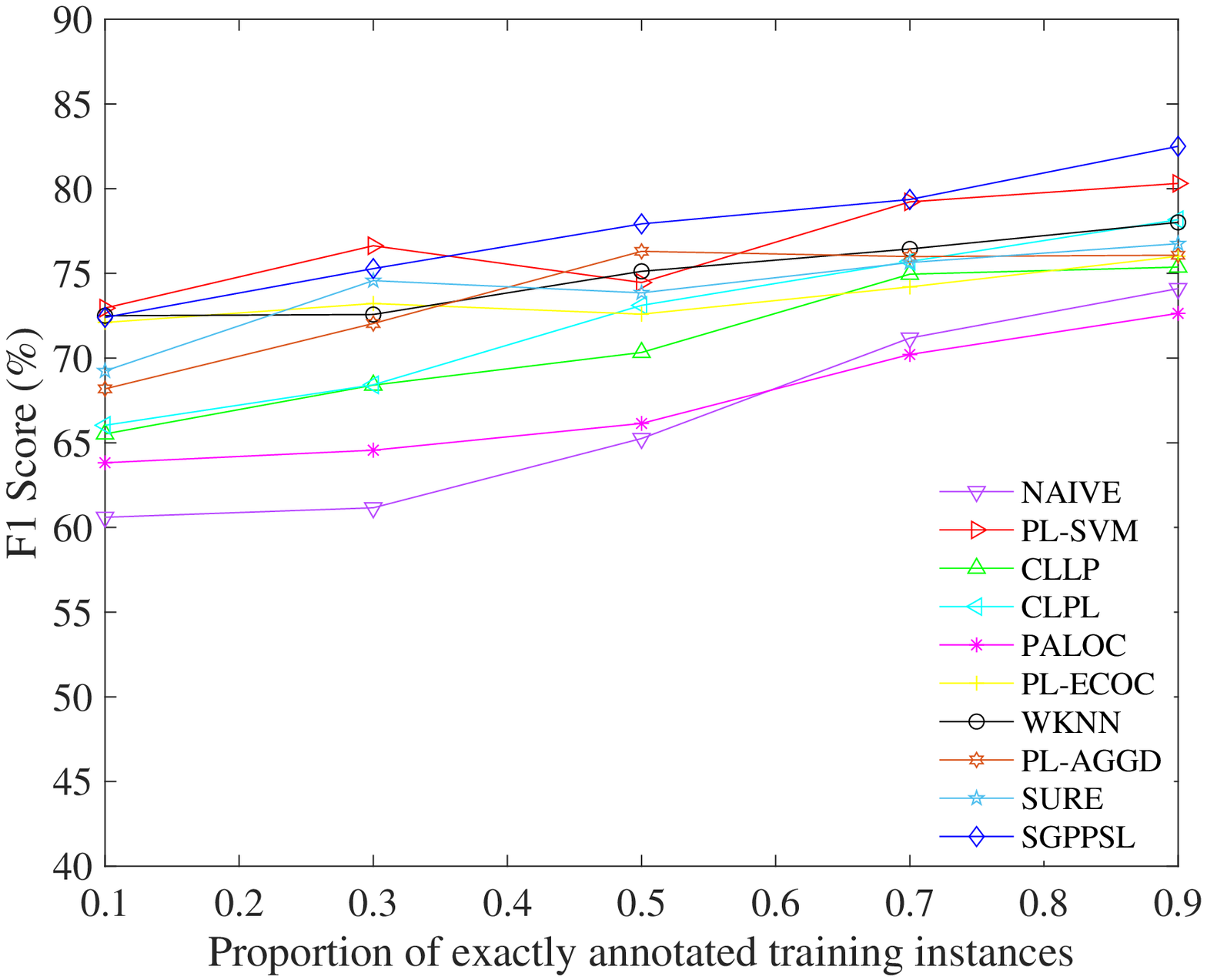}
  \subcaption{$cl$ = 3}\label{fig:awesome_image2}
\endminipage\hfill
\minipage{0.32\textwidth}%
  \includegraphics[width=\linewidth]{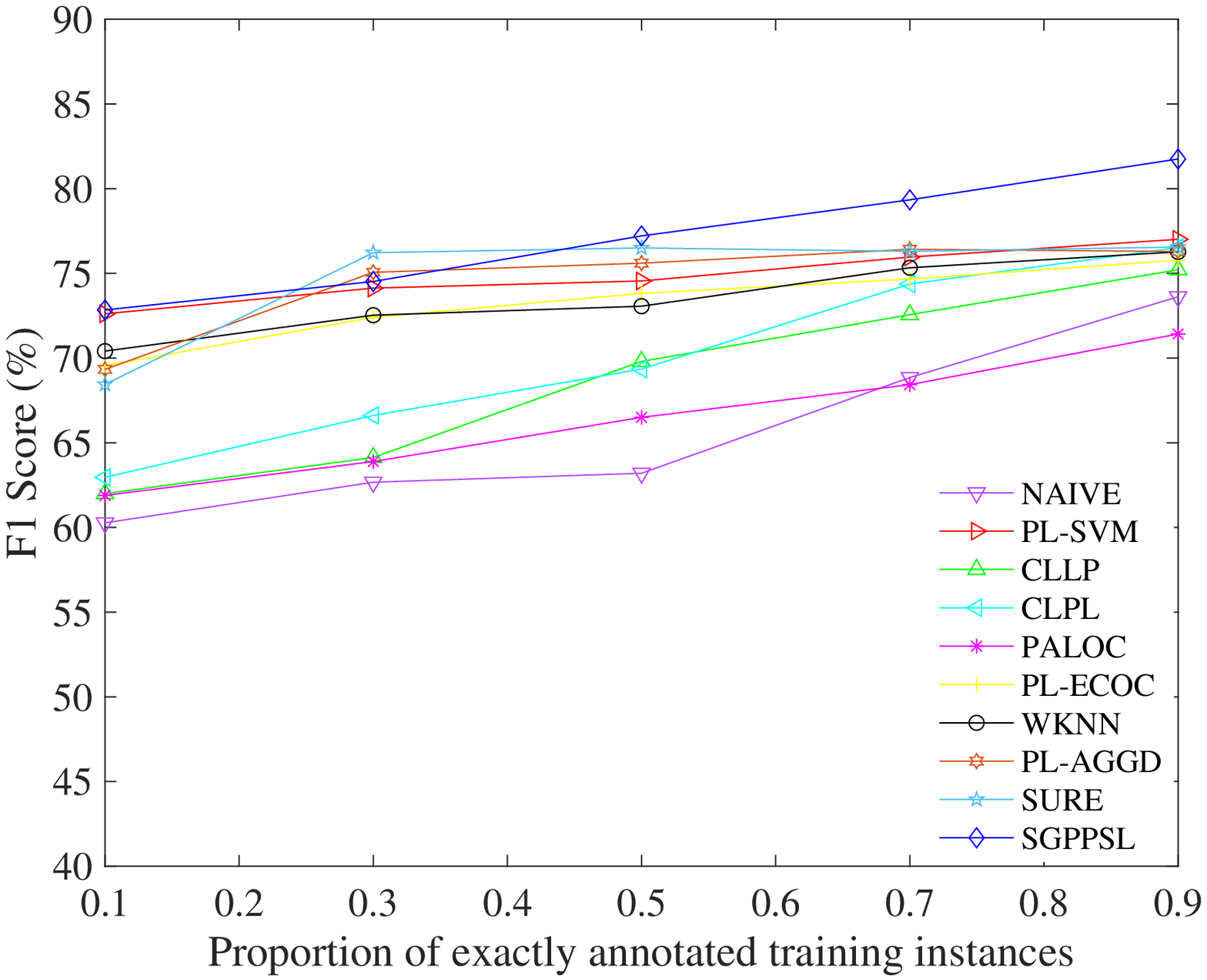}
  \subcaption{$cl$ = 4}\label{fig:awesome_image3}
\endminipage
\caption{The performance of Base NP  with varying $cl$.}
\end{figure*}

\iffalse
\begin{figure}[h]
\centering
\includegraphics[width=3.5in,height = 2.3in ]{np1.eps}
\caption {The performance of Base NP ($cl$=2).}
\label{fig:secondfigure}
\end{figure}

\begin{figure}
\centering
\includegraphics[width=3.5in,height = 2.3in ]{np2.eps}
\caption {The performance of Base NP ($cl$=3).}
\label{fig:secondfigure}
\end{figure}

\begin{figure}
\centering
\includegraphics[width=3.5in,height = 2.3in ]{np3.eps}
\caption {The performance of Base NP ($cl$=4).}
\label{fig:secondfigure}
\end{figure}
\fi

2) Chunking.

Figure 7 demonstrates the performance of compared methods on the Chunking task. It is obvious that SGPPSL consistently outperforms the other compared methods. By increasing the proportion of annotated training instances, SGPPSL, PL-SVM and CLPL performs more stably than the other baselines. When $cl = 4$, the performance of PSL based models is positively correlated with the proportion of exactly annotated sequences. Moreover, with the increase of candidate labels, the performance with lower $p$ is decreased.

Different from Base NP task, the size of label set for Chunking is relatively large (as described in Table \Romannum{1}). SGPPSL addresses different contributions of candidate labels in the learning process, which not only avoids wrong assignment of the ground-truth but takes the confusing label information with inherent ambiguity of language into account. For example, ``holding" in the sentence ``...seven directors of each holding company..." has three candidate labels $\left \{[``\textbf{I-NP}", ``I-VP", ``B-VP"] \right \}$ (ground-truth is highlighted in bold). As `holding" can be ``B-VP" in the sentence ``...there is little holding sterling firm...", learning the specific weight to the pair [``holding" $\rightarrow$ ``B-VP"] also expresses how confusing the label pair [``I-NP", ``B-VP"] to the word ``holding", which can help improve the performance of the model in identifying the similar word as ``I-NP" or ``B-VP".

\begin{figure*}[!htb]
\minipage{0.32\textwidth}
  \includegraphics[width=\linewidth]{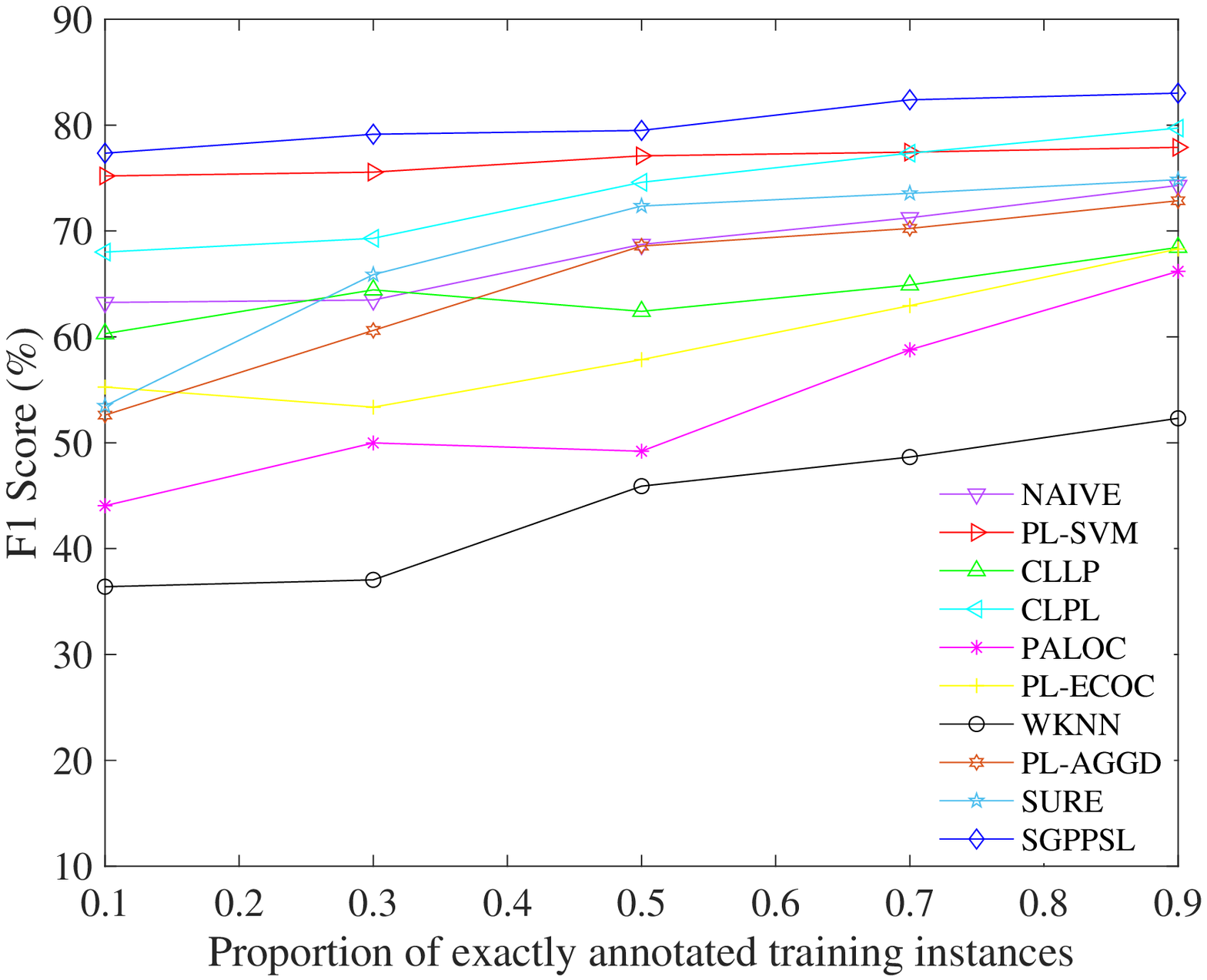}
  \subcaption{$cl$ = 2}\label{fig:awesome_image1}
\endminipage\hfill
\minipage{0.32\textwidth}
  \includegraphics[width=\linewidth]{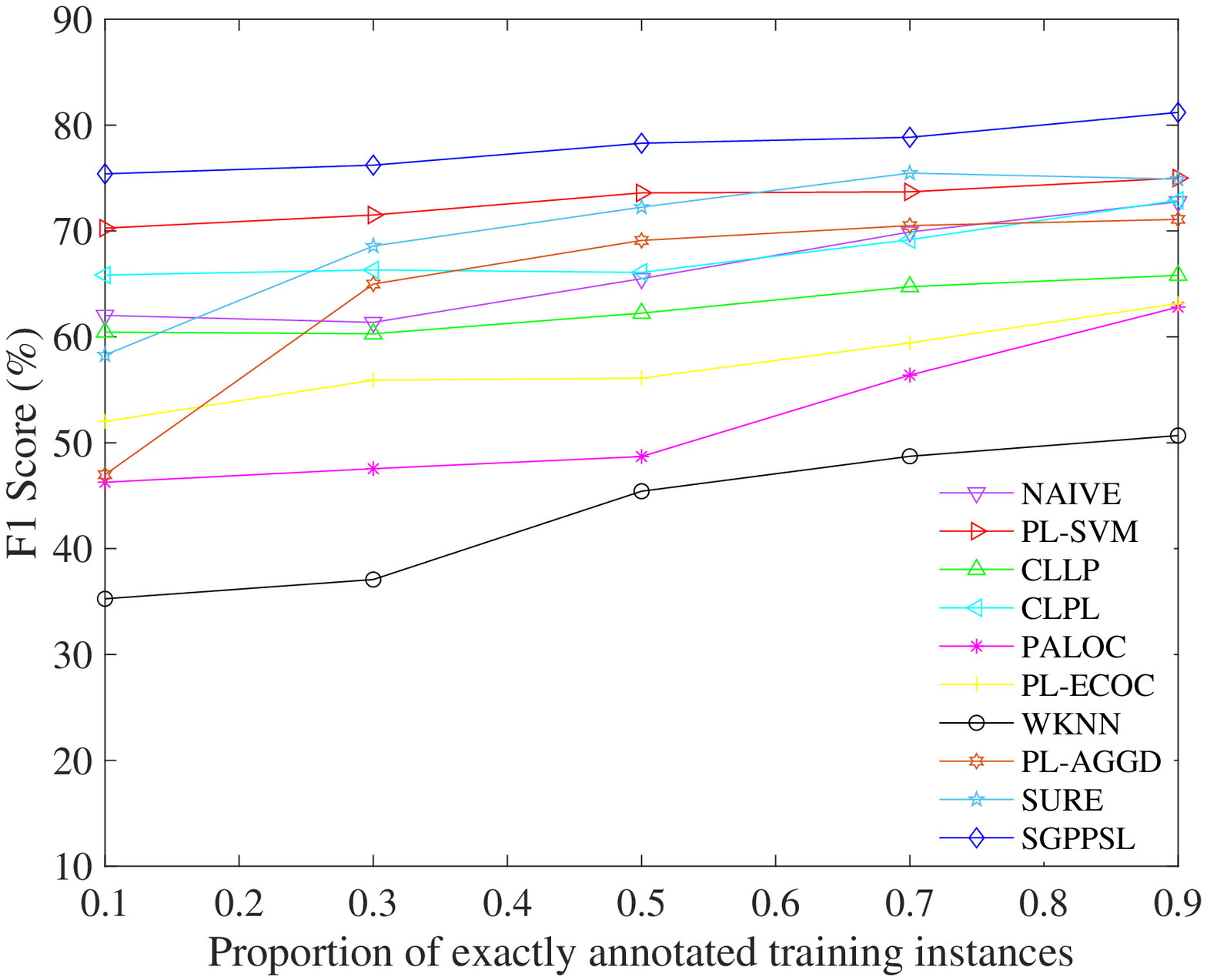}
  \subcaption{$cl$ = 3}\label{fig:awesome_image2}
\endminipage\hfill
\minipage{0.32\textwidth}%
  \includegraphics[width=\linewidth]{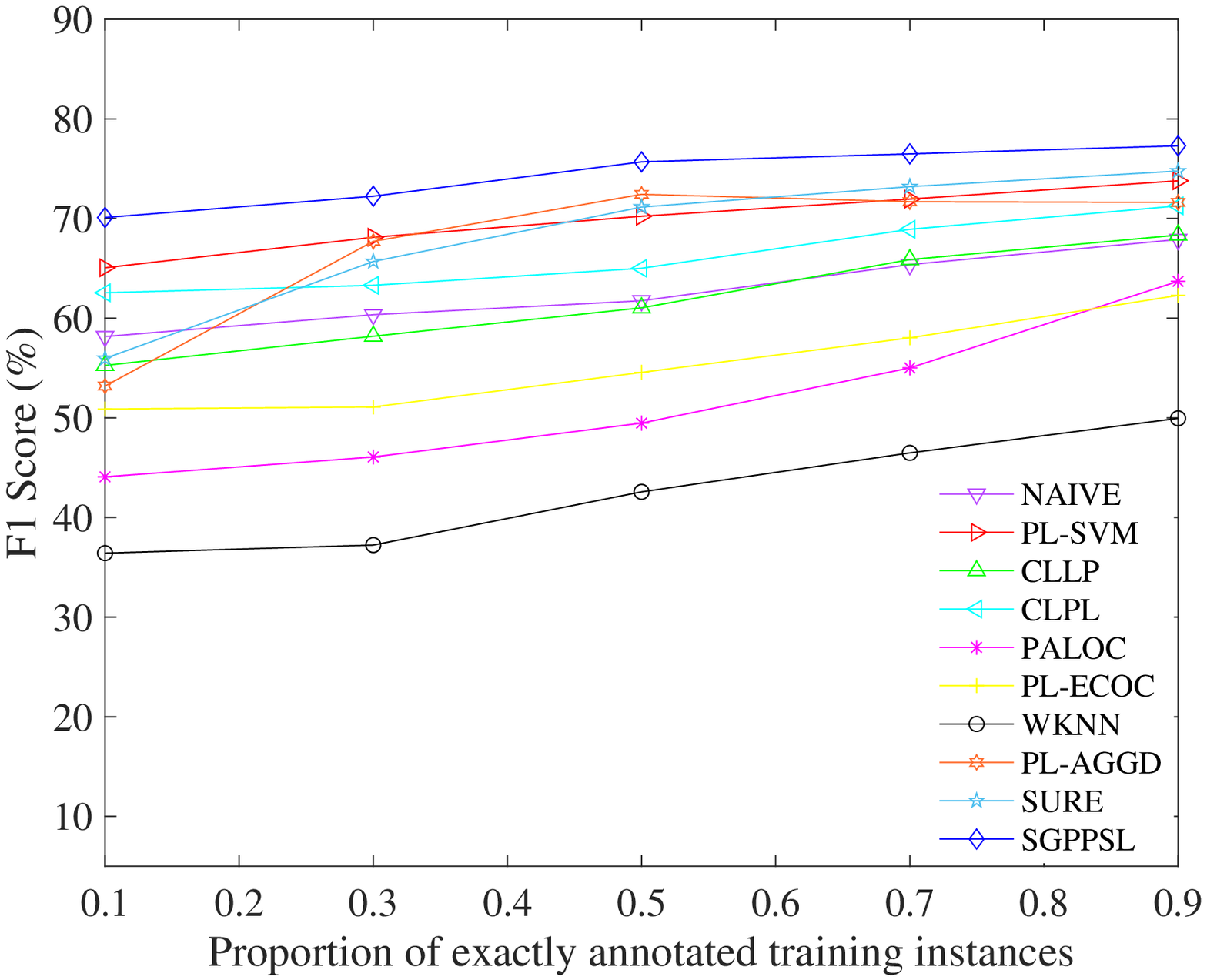}
  \subcaption{$cl$ = 4}\label{fig:awesome_image3}
\endminipage
\caption{The performance of Chunking with varying $cl$.}
\end{figure*}

\iffalse
\begin{figure}[h]
\centering
\includegraphics[width=3.5in,height = 2.3in ]{ck1.eps}
\caption {The performance of Chunking ($cl$=2).}
\label{fig:secondfigure}
\end{figure}

\begin{figure}
\centering
\includegraphics[width=3.5in,height = 2.3in ]{ck2.eps}
\caption {The performance of Chunking ($cl$=3).}
\label{fig:secondfigure}
\end{figure}

\begin{figure}
\centering
\includegraphics[width=3.5in,height = 2.3in ]{ck3.eps}
\caption {The performance of Chunking ($cl$=4).}
\label{fig:secondfigure}
\end{figure}
\fi

3) NER.

Figure 8 presents the results on NER task. It can be seen that SGPPSL outperforms the other compared methods when $p < 0.7$. CLPL show its superiority with the very close results compared with that of SGPPSL. When $cl = 4$, these PSL based methods can achieve better results by increasing the proportion of exactly annotated sequences. 

There is also confusing label information in NER task. For example, ``European" is tagged as ``B-MISC" in the sentence ``...of their European Champions Cup..."  while ``B-ORG" in the sentence ``...membership of the European Union...". Although the size of label set for NER is relatively small, confidence weighted mechanism enables SGPPSL achieves better performance with lower $p$.

\begin{figure*}[!htb]
\minipage{0.32\textwidth}
  \includegraphics[width=\linewidth]{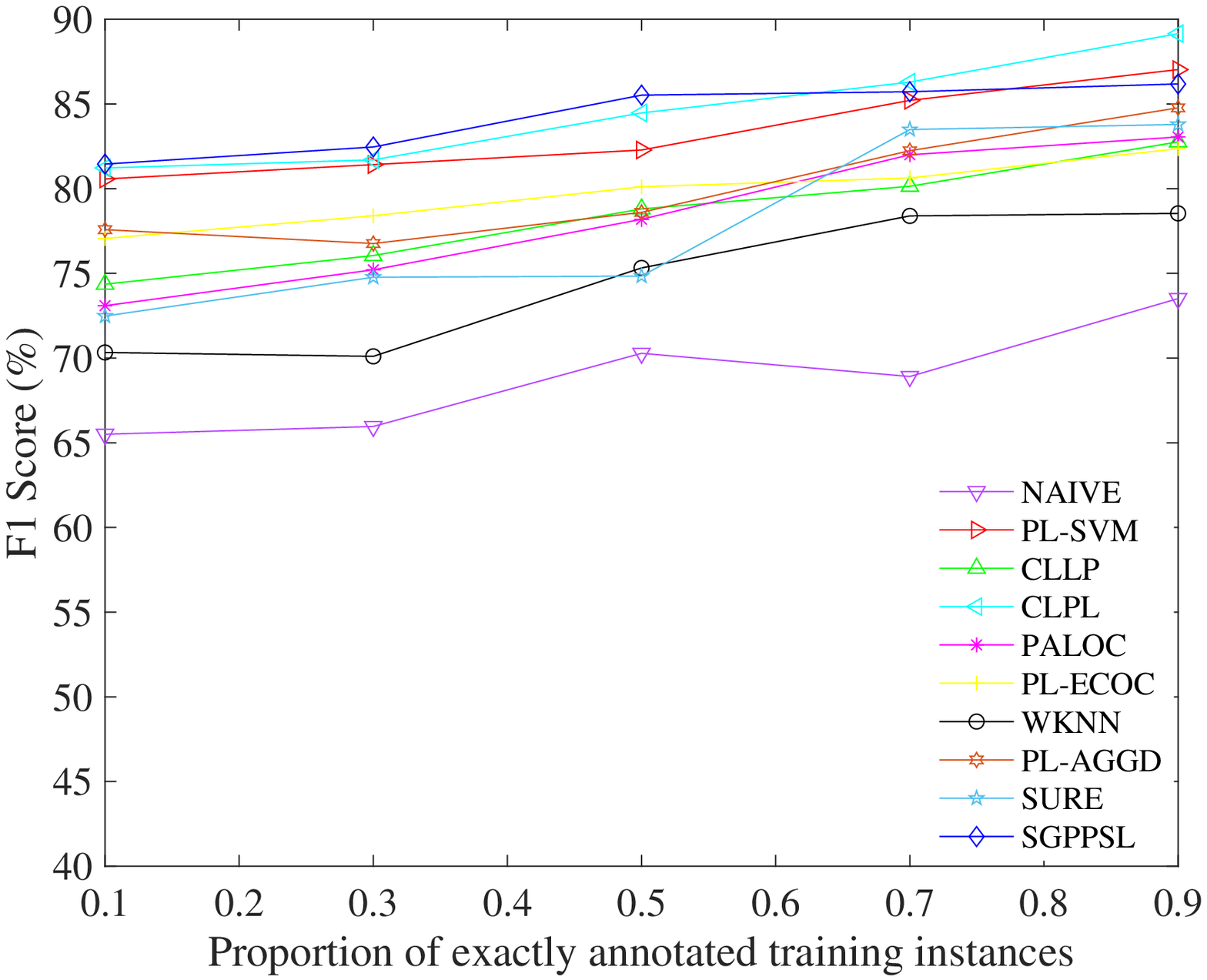}
  \subcaption{$cl$ = 2}\label{fig:awesome_image1}
\endminipage\hfill
\minipage{0.32\textwidth}
  \includegraphics[width=\linewidth]{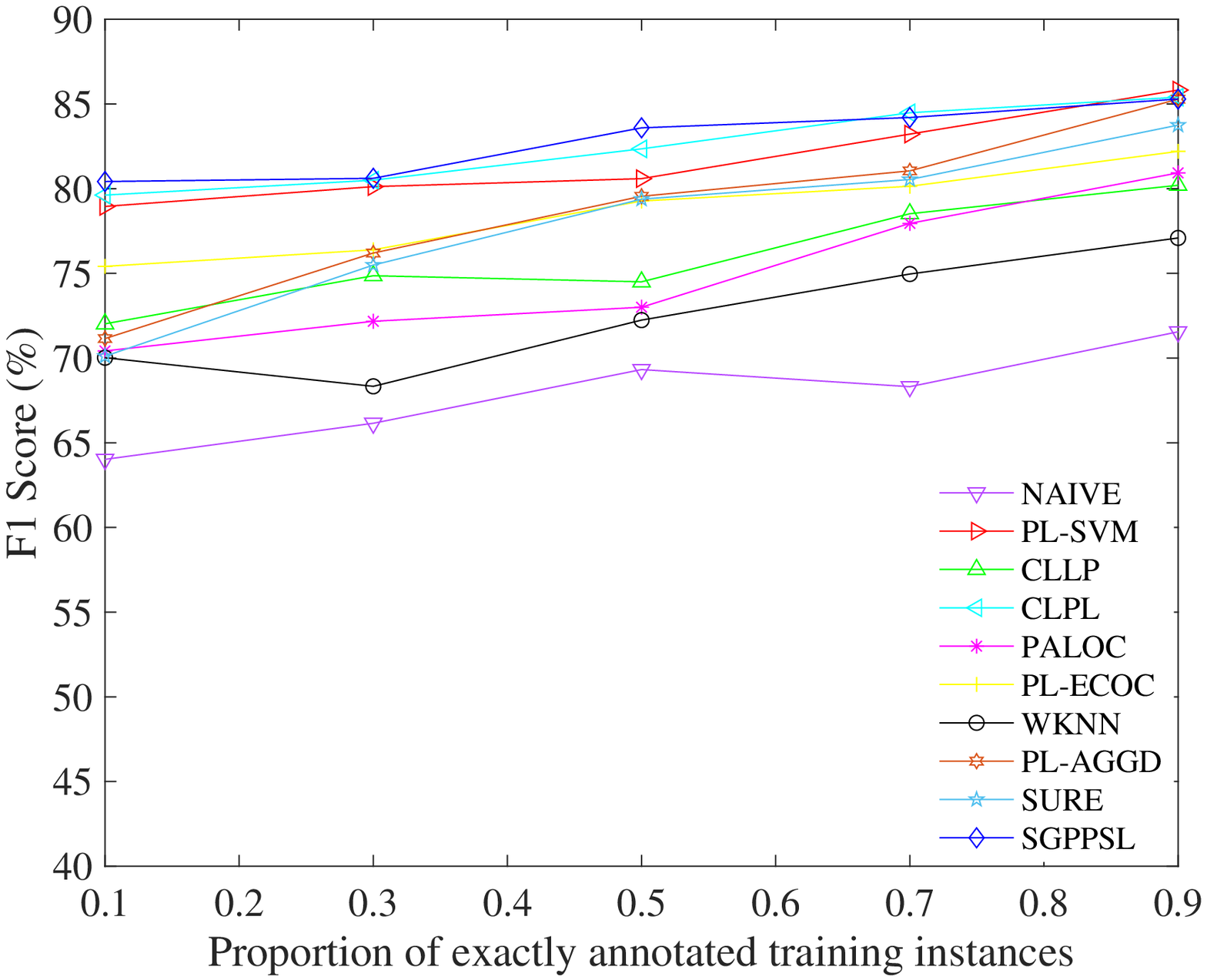}
  \subcaption{$cl$ = 3}\label{fig:awesome_image2}
\endminipage\hfill
\minipage{0.32\textwidth}%
  \includegraphics[width=\linewidth]{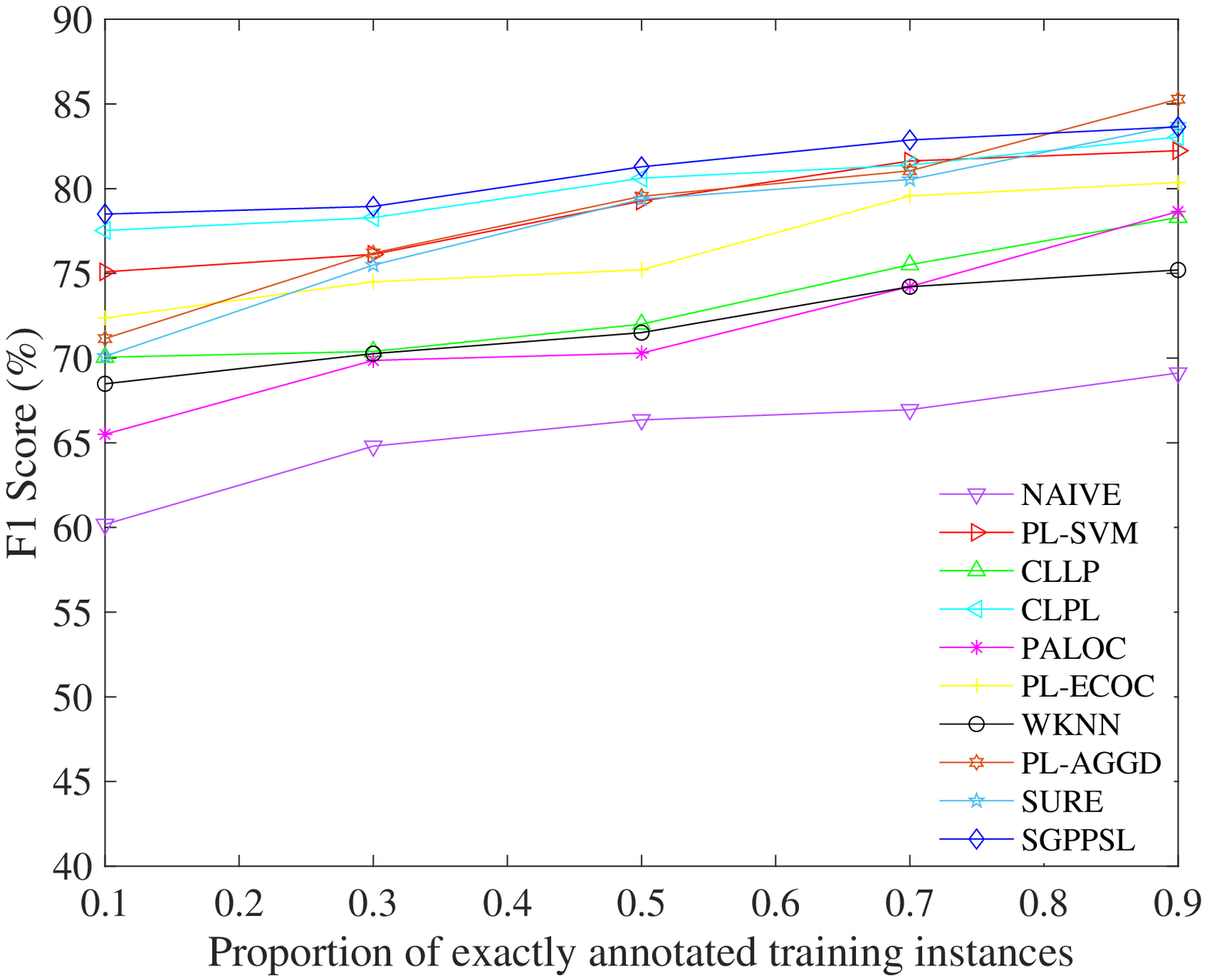}
  \subcaption{$cl$ = 4}\label{fig:awesome_image3}
\endminipage
\caption{The performance of NER with varying $cl$.}
\end{figure*}

\iffalse
\begin{figure}[h]
\centering
\includegraphics[width=3.5in,height = 2.3in ]{ner1.eps}
\caption {The performance of NER ($cl$=2).}
\label{fig:secondfigure}
\end{figure}

\begin{figure}
\centering
\includegraphics[width=3.5in,height = 2.3in ]{ner2.eps}
\caption {The performance of NER ($cl$=3).}
\label{fig:secondfigure}
\end{figure}

\begin{figure}
\centering
\includegraphics[width=3.5in,height = 2.3in ]{ner3.eps}
\caption {The performance of NER ($cl$=4).}
\label{fig:secondfigure}
\end{figure}
\fi

Based on the above observations on the performance of three tasks, SGPPSL is more effective in handling partial annotations with much confusing label information. While PL-SVM and CLPL obtain competitive results in some settings, average and unique disambiguation strategy may ignore the contribution of the ground-truth label and inherent ambiguity in the language, which may adversely affects the performance of NLP tasks. Generally increasing the proportion of exactly annotated instance can help uncover the ground-truth labels and thus enhance the performance.

\subsection{Analyzing different sampling toward generating candidate label set}
In this paper, we generate candidate label set with size $cl$ by conducting random sampling $cl$ times and each randomly select a negative label to the candidates. To enable different label partialities among partially labeled training samples, we employ the sampling strategy as described in \cite{re38} by controlling the flipping probability $r$ that a negative label is flipped to candidate label set. For this sampling strategy, we choose less-partial $r=0.05$ and strong-partial $r=0.75$ scenarios with $p=0.5$. Also, to give a more intuitive comparison, we report the results of $cl = 2$ with $p=0.5$ presented in Section \Romannum{4} D.

Figure 9 demonstrates the performance with different sampling toward generating candidate label set. We can see that in most cases the results of $r=0.05$ are consistent with the results of $cl=2$ as in less-partial setting most of partially annotated samples have only one additional negative label. While in strong-partial setting, it is more likely to assign around $\left |  \mathcal{Y}\right |$ negative labels to the candidate label set, which inevitably results in lower performance compared with the results of $r=0.05$.  Furthermore, as the label space is limited in Base NP (i.e. $\left |  \mathcal{Y} \right |=3$), the performance with $r=0.75$ approximates to the results of $cl=3$.

\begin{figure*}[!htb]
\minipage{0.32\textwidth}
  \includegraphics[width=\linewidth]{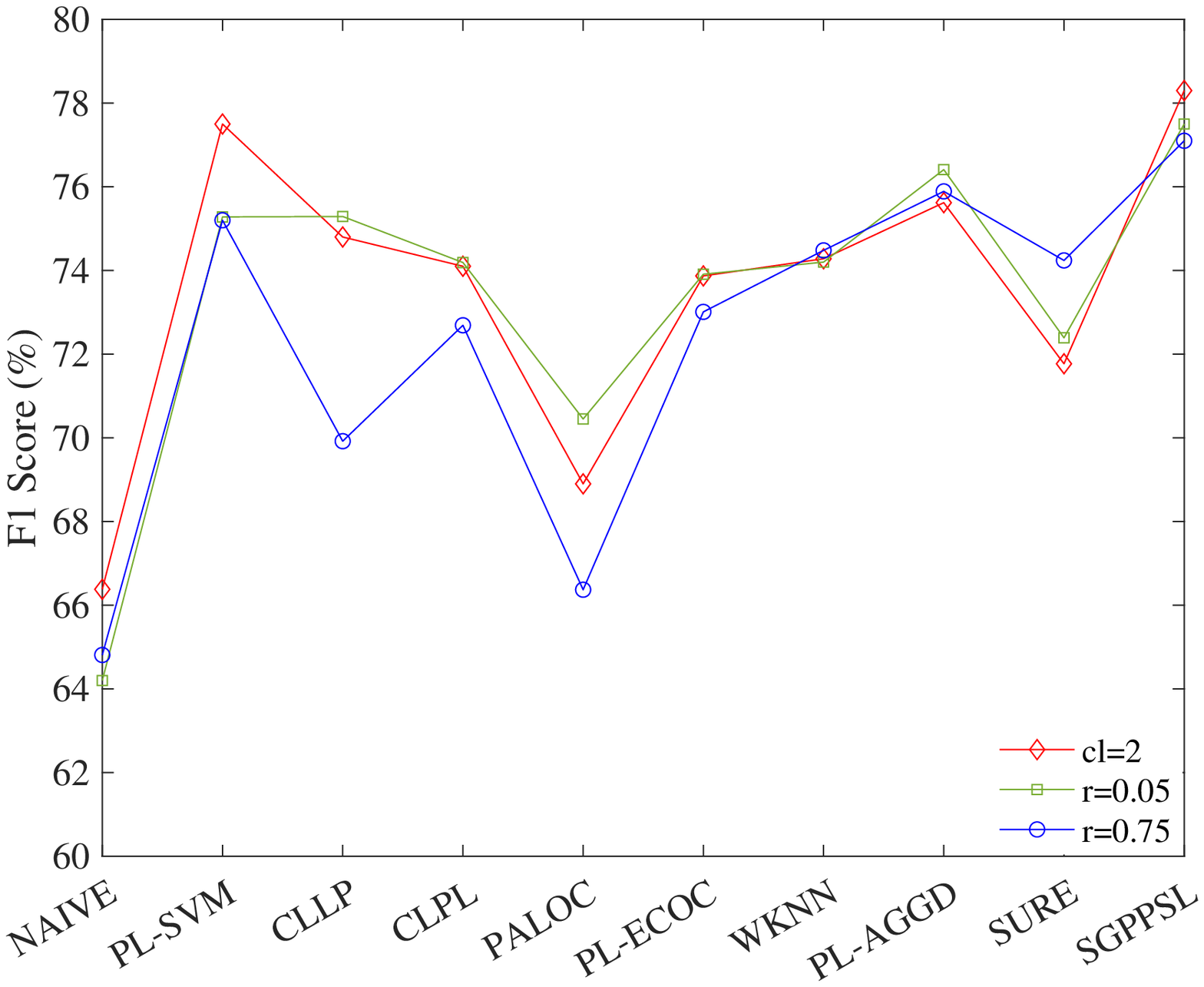}
  \subcaption{Base NP}\label{fig:awesome_image1}
\endminipage\hfill
\minipage{0.32\textwidth}
  \includegraphics[width=\linewidth]{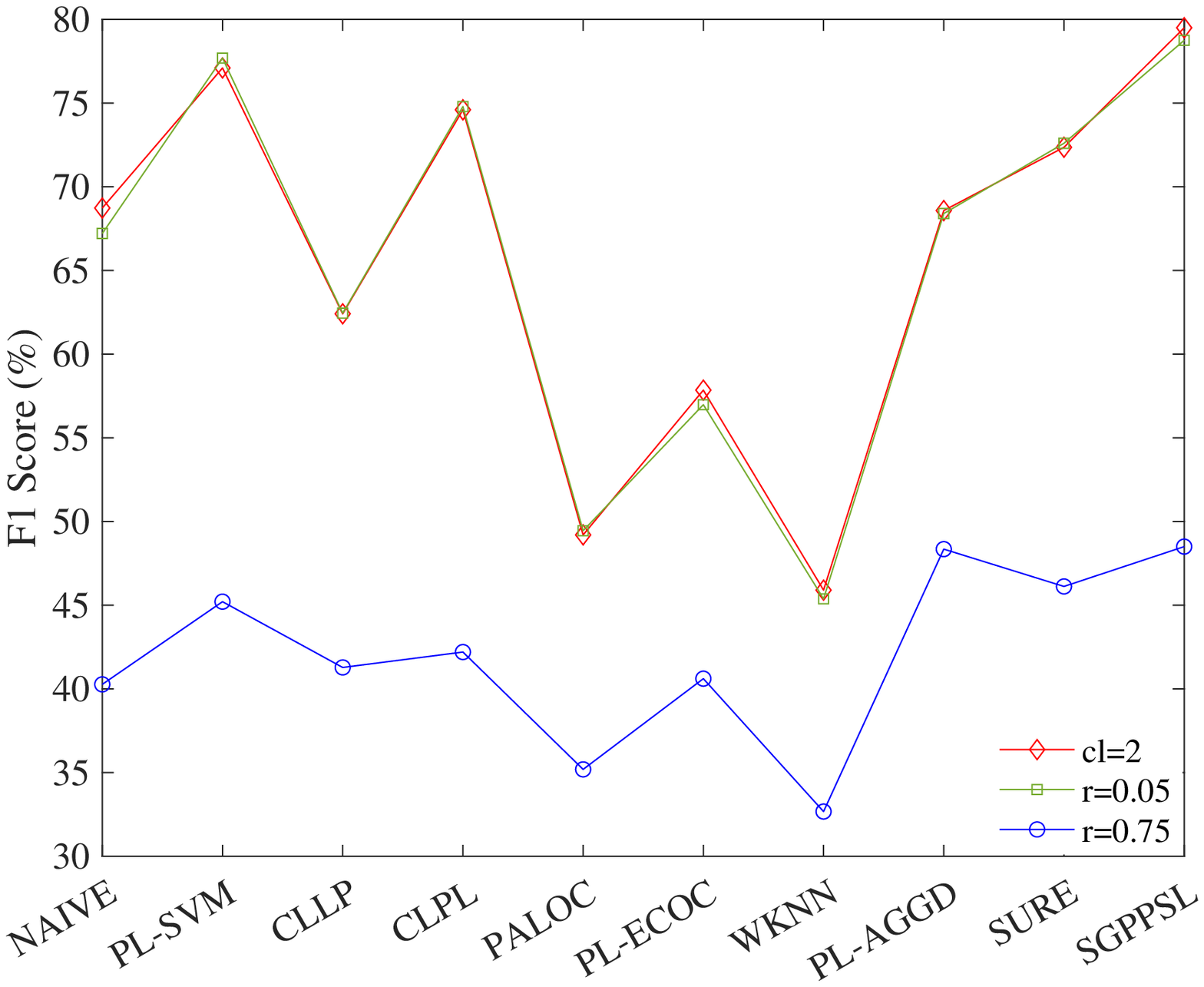}
  \subcaption{Chunking}\label{fig:awesome_image2}
\endminipage\hfill
\minipage{0.32\textwidth}%
  \includegraphics[width=\linewidth]{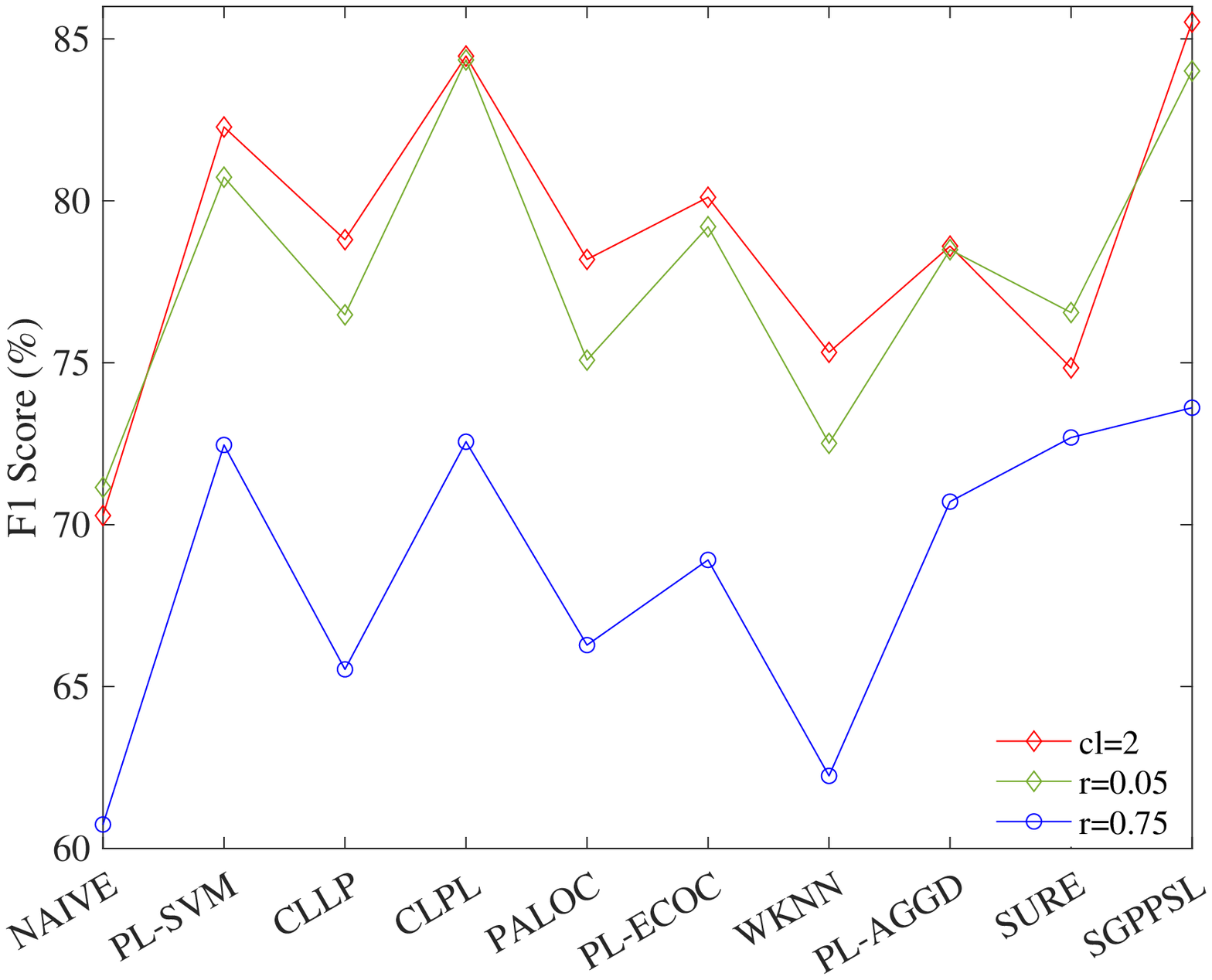}
  \subcaption{NER}\label{fig:awesome_image3}
\endminipage
\caption{The performance with different sampling toward generating candidate label set.}
\end{figure*}

\subsection{Evaluating prediction algorithms}
To verify the proposed weighted Viterbi algorithm for prediction, we compare it with the traditional Viterbi, where $cl=3$ as the SGPPSL performs stably with varying $cl$. Table \Romannum{3} report the performance of different prediction algorithms. As shown in Table  \Romannum{3} , the performance of both decoding algorithms is positively correlated with the proportion of exactly annotated instances. Furthermore, in most cases weighted Viterbi outperforms Viterbi, especially with lower $p$. 

With the large proportion of ambiguous annotations, the uncertainty of label assignment in prediction is greatly increased. The weighted Viterbi algorithm incorporates confidence measures estimated in the training process to reduce the uncertainly arose from ambiguous annotations. For example, as shown in Figure 3, supposing the set of ambiguous annotations of the neighborhoods of the word ``lead" is [``NN",``NNP",``VBD",``VBP"] when assigning the POS tag to ``lead" in the decoding, the ground-truth label ``NN" obtains the highest score as in most cases our proposed model can identify the ground-truth with the greatest confidence measure.

\renewcommand{\arraystretch}{1.5}
\begin{table}

\centering
\caption{The performance of comparing prediction algorithms (\%)}
\scalebox{0.9}{
\begin{tabular}{llccccc} 
\hline
Task                      & Prediction algorithm & 0.1            & 0.3            & 0.5            & 0.7            & 0.9             \\ 
\hline
\multirow{2}{*}{Base NP~} & Viterbi              & 69.53        & 71.80         & 76.18       & \textbf{80.62}         & \textbf{82.95}     \\
                          & Weighted Viterbi     & \textbf{72.40} & \textbf{75.28} & \textbf{77.92} & 79.36              & 82.50  \\ 
\hline
\multirow{2}{*}{Chunking} & Viterbi              & 69.30          & 70.54          & 74.69          & \textbf{79.24} & 80.37           \\
                          & Weighted Viterbi     & \textbf{75.40} & \textbf{76.22} & \textbf{78.29} & 78.85          & \textbf{81.20}  \\ 
\hline
\multirow{2}{*}{NER}      & Viterbi              & 72.91          & 75.70          & 81.35          & \textbf{85.87} & \textbf{85.96}  \\
                          & Weighted Viterbi     & \textbf{80.42} & \textbf{80.61} & \textbf{83.59} & 84.20          & 85.29           \\
\hline
\end{tabular}}
\end{table}

\section{Conclusion}
In this paper, we propose a non-parametric Bayesian model SGPPSL for partial sequence labeling. The proposed SGPPSL employs factor-as-piece likelihood approximation and confidence measure for each candidate label, which effectively avoids handling large number of candidate output sequences generated by partially annotated data and addresses the different contribution of each candidate label. A weighted Viterbi algorithm is proposed to incorporate confidence measure in prediction. We conduct the experiments on the tasks Base NP, Chunking and NER. The experimental results show that the proposed SGPPSL is more effective in handling partial annotations with much confusing label information. Furthermore, the weighted Viterbi achieves better performance than the traditional Viterbi. In the future, we will consider efficient variational inference for structured Gaussian Processes.

\section{Acknowledgements}
This work was fully supported by the Hong Kong Government GRF under project CityU 11216620.
\bibliographystyle{ieeetr}
\bibliography{ref}

\end{document}